\ifdefined\pdftexversion\pdfoutput=1\fi  
\documentclass{article}

\usepackage[preprint]{neurips_2025}

\usepackage[utf8]{inputenc}
\usepackage[T1]{fontenc}
\usepackage{hyperref}
\usepackage{url}
\usepackage{booktabs}
\usepackage{amsmath,amssymb,amsfonts}
\usepackage{nicefrac}
\usepackage{microtype}
\usepackage{xcolor}
\usepackage{graphicx}
\usepackage{textcomp}
\usepackage{multirow}
\usepackage{makecell}
\usepackage{float}
\usepackage{caption}
\usepackage{subcaption}
\usepackage{enumitem}
\usepackage{array}

\hypersetup{colorlinks=true,linkcolor=blue!50!black,citecolor=blue!50!black,urlcolor=blue!50!black}

\makeatletter
\let\bg@textminus\textminus
\DeclareRobustCommand{\textminus}{\ifmmode -\else\bg@textminus\fi}
\makeatother

\newcommand{\ci}[2]{\ifmmode\text{\scriptsize\,[#1,\,#2]}\else{\scriptsize\,[#1,\,#2]}\fi}

\newlength{\tblsep}
\newcommand{\afail}{a^{\times}}
\newcommand{\agold}{a^{\checkmark}}
\newcommand{\Gain}{G}
\newcommand{\copyterm}{\Delta_{\mathrm{copy}}}
\newcommand{\semterm}{\Delta_{\mathrm{sem}}}
\newcommand{\ctxpre}{C}

\newcommand{\mv}[1]{\texttt{#1}}   
\newcommand{\RepoURL}{\url{https://github.com/Esmail-ibraheem/feedback-that-backfires/tree/games}}

\newcommand{\NumPointsSokoban}{100}
\newcommand{\NumPointsConnectFour}{80}
\newcommand{\NumPointsOthello}{80}

\newcommand{\PerTypeQuota}{20}
\newcommand{\NRandomItems}{20}

\newcommand{\SokobanMaxPrefix}{3}
\newcommand{\FailedInPrefixConnectFour}{15/80}
\newcommand{\FailedInPrefixSokoban}{24/100}

\newcommand{\ModelListRQone}{SmolLM2-135M, SmolLM2-360M, Qwen2.5-0.5B, Qwen3-0.6B, Llama-3.2-1B, Qwen2.5-1.5B, SmolLM2-1.7B, Qwen3-1.7B}
\newcommand{\NumModelsRQone}{8}
\newcommand{\SmallestModelRQone}{135M}
\newcommand{\LargestModelRQone}{1.7B}

\newcommand{\RegressionG}{\textminus 23.92}
\newcommand{\PaperGOriginal}{\textminus 23.92}
\newcommand{\RegressionMaxDiff}{$9.5\times 10^{-5}$}
\newcommand{\RegressionCopy}{21.69}
\newcommand{\RegressionSem}{2.23}
\newcommand{\RegressionGPUG}{\textminus 23.92}
\newcommand{\RegressionGPUMaxDiff}{$3.5\times 10^{\textminus 4}$}
\newcommand{\RegressionGPUTol}{$10^{\textminus 2}$}
\newcommand{\CPUName}{Intel(R) Core(TM) i7-6820HQ CPU @ 2.70GHz}
\newcommand{\CPUTorchVersion}{2.4.0+cpu}
\newcommand{\GPUName}{Tesla T4}
\newcommand{\TorchVersion}{2.11.0+cu130}
\newcommand{\CudaVersion}{13.0}
\newcommand{\TransformersVersion}{4.56.1}
\newcommand{\PythonVersion}{3.13.15}
\newcommand{\TFThirtyTwo}{disabled}
\newcommand{\NumProbeRecords}{51{,}726}
\newcommand{\RQoneG}{\textminus 1.38}
\newcommand{\RQoneGCI}{\textminus 1.38\,\ci{\textminus 1.85}{\textminus 0.95}}
\newcommand{\RQoneCopy}{1.34}
\newcommand{\RQoneCopyCI}{+1.34\,\ci{+0.95}{+1.76}}
\newcommand{\RQoneSem}{0.04}
\newcommand{\RQoneSemCI}{+0.04\,\ci{\textminus 0.10}{+0.17}}
\newcommand{\RQoneNItems}{840}
\newcommand{\RQoneNPositions}{120}
\newcommand{\RQoneFracNeg}{61\%}
\newcommand{\RQonePRepeatPre}{0.24}
\newcommand{\RQonePRepeatPost}{0.35}
\newcommand{\RQoneGreedyPre}{12\%}
\newcommand{\RQoneGreedyPost}{23\%}

\newcommand{\RQoneIllegalGCI}{\textminus 1.61\,\ci{\textminus 2.16}{\textminus 1.11}}

\newcommand{\RQoneIllegalCopyCI}{+1.52\,\ci{+1.07}{+2.01}}

\newcommand{\RQoneIllegalFracNeg}{66\%}

\newcommand{\RQoneLosingGCI}{\textminus 0.39\,\ci{\textminus 0.90}{+0.06}}

\newcommand{\RQoneLosingCopyCI}{+0.59\,\ci{+0.10}{+1.09}}

\newcommand{\RQoneLosingNPositions}{23}

\newcommand{\RQoneLosingGreedyPre}{26\%}
\newcommand{\RQoneLosingGreedyPost}{21\%}

\newcommand{\RQoneLosingGSokobanCI}{+0.23\,\ci{\textminus 0.26}{+0.60}}

\newcommand{\RQoneLosingGConnectFourCI}{\textminus 0.84\,\ci{\textminus 1.48}{\textminus 0.22}}

\newcommand{\RQoneGQwenThreeSCI}{\textminus 5.45\,\ci{\textminus 6.08}{\textminus 4.85}}

\newcommand{\RQoneGLlamaCI}{+0.66\,\ci{+0.26}{+1.05}}

\newcommand{\RQoneCopyLlamaCI}{\textminus 0.52\,\ci{\textminus 0.95}{\textminus 0.08}}

\newcommand{\RQoneGQwenTwoSSokobanCI}{\textminus 0.33\,\ci{\textminus 1.54}{+0.85}}

\newcommand{\RQoneNumNegPoint}{7}
\newcommand{\RQoneNumNeg}{4}

\newcommand{\RQoneNumNull}{3}
\newcommand{\RQoneNegList}{SmolLM2-135M, SmolLM2-360M, Qwen3-0.6B and SmolLM2-1.7B}
\newcommand{\RQonePosList}{Llama-3.2-1B}
\newcommand{\RQoneNullList}{Qwen2.5-0.5B, Qwen2.5-1.5B and Qwen3-1.7B}

\newcommand{\RQoneModelSrcGCI}{+1.34\,\ci{+1.18}{+1.49}}
\newcommand{\RQoneModelSrcFracNeg}{26\%}
\newcommand{\RQoneModelSrcNItems}{759}

\newcommand{\RQtwoGBaseCI}{\textminus 0.52\,\ci{\textminus 0.70}{\textminus 0.34}}

\newcommand{\RQtwoCopyBaseCI}{+0.49\,\ci{+0.31}{+0.67}}

\newcommand{\RQtwoSemBaseCI}{+0.03\,\ci{+0.02}{+0.05}}
\newcommand{\RQtwoFracNegBase}{82\%}

\newcommand{\RQtwoGDPOCI}{\textminus 0.12\,\ci{\textminus 0.39}{+0.15}}

\newcommand{\RQtwoCopyDPOCI}{\textminus 0.23\,\ci{\textminus 0.51}{+0.05}}

\newcommand{\RQtwoSemDPOCI}{+0.35\,\ci{+0.30}{+0.39}}

\newcommand{\RQtwoNItems}{180}
\newcommand{\RQtwoNPositions}{60}

\newcommand{\RQtwoDGBaseSFTCI}{+0.10\,\ci{\textminus 0.00}{+0.20}}

\newcommand{\RQtwoDCopyBaseSFTCI}{\textminus 0.21\,\ci{\textminus 0.32}{\textminus 0.09}}

\newcommand{\RQtwoDGSFTDPOCI}{+0.30\,\ci{+0.21}{+0.38}}

\newcommand{\RQtwoDCopySFTDPOCI}{\textminus 0.51\,\ci{\textminus 0.59}{\textminus 0.42}}

\newcommand{\RQtwoDSemSFTDPOCI}{+0.21\,\ci{+0.18}{+0.24}}

\newcommand{\RQtwoDGDPORLVRCI}{\textminus 0.03\,\ci{\textminus 0.06}{\textminus 0.00}}

\newcommand{\RQtwoDGRLVRInstructCI}{\textminus 0.01\,\ci{\textminus 0.03}{+0.01}}

\newcommand{\RQtwoChatDGSFTDPOCI}{+0.46\,\ci{+0.33}{+0.59}}

\newcommand{\MechRQtwoVerbose}{$+0.48$ (SE $0.25$, $p=0.056$)}
\newcommand{\MechRQtwoIndex}{$\textminus 0.19$ (SE $0.08$, $p=0.015$)}
\newcommand{\MechRQtwoLength}{$\textminus 0.36$ (SE $0.11$, $p<0.001$)}
\newcommand{\MechRQtwoOverlap}{$+0.86$ (SE $0.17$, $p<0.001$)}

\newcommand{\MechRQoneVerbose}{$\textminus 0.16$ (SE $0.78$, $p=0.835$)}
\newcommand{\MechRQoneIndex}{$\textminus 0.06$ (SE $0.20$, $p=0.756$)}
\newcommand{\MechRQoneLength}{$\textminus 0.03$ (SE $0.28$, $p=0.908$)}
\newcommand{\MechRQoneOverlap}{$+1.12$ (SE $0.36$, $p=0.002$)}

\newcommand{\MechRQtwoPriorOverlap}{$\textminus 0.55$ (SE $0.11$, $p<0.001$)}
\newcommand{\MechRQtwoPriorPrior}{$+0.81$ (SE $0.02$, $p<0.001$)}

\newcommand{\MechRQonePriorVerbose}{$\textminus 2.12$ (SE $0.34$, $p<0.001$)}

\newcommand{\MechRQonePriorPrior}{$+0.92$ (SE $0.01$, $p<0.001$)}

\newcommand{\RQthreeIllegalGQwenTwoSCI}{\textminus 2.08\,\ci{\textminus 3.24}{\textminus 0.97}}

\newcommand{\RQthreeIllegalCopyQwenTwoSCI}{+2.05\,\ci{+0.75}{+3.36}}

\newcommand{\RQthreeNativeGQwenTwoSCI}{\textminus 2.39\,\ci{\textminus 3.75}{\textminus 1.02}}

\newcommand{\RQthreeLosingGQwenTwoSCI}{+0.63\,\ci{\textminus 0.26}{+1.47}}
\newcommand{\RQthreeIllegalNQwenTwoS}{45}
\newcommand{\RQthreeLosingNQwenTwoS}{15}

\newcommand{\RQthreeIllegalGQwenTwoLCI}{\textminus 3.82\,\ci{\textminus 5.54}{\textminus 2.22}}

\newcommand{\RQthreeIllegalCopyQwenTwoLCI}{+3.87\,\ci{+2.29}{+5.54}}

\newcommand{\RQthreeNativeGQwenTwoLCI}{\textminus 3.89\,\ci{\textminus 5.52}{\textminus 2.33}}

\newcommand{\RQthreeLosingGQwenTwoLCI}{+0.50\,\ci{+0.29}{+0.75}}

\newcommand{\RQfourAbstractCI}{\textminus 0.83\,\ci{\textminus 1.08}{\textminus 0.58}}

\newcommand{\RQfourAbstractMinCI}{\textminus 0.72\,\ci{\textminus 0.97}{\textminus 0.47}}

\newcommand{\RQfourInstrCI}{+0.03\,\ci{\textminus 0.00}{+0.06}}

\newcommand{\RQfourEchoCI}{+0.78\,\ci{+0.70}{+0.86}}

\newcommand{\RQfourTwiceCI}{+1.68\,\ci{+1.51}{+1.86}}

\newcommand{\RQfourOtherCI}{\textminus 0.77\,\ci{\textminus 1.02}{\textminus 0.54}}
\newcommand{\RQfourNumModels}{3}
\newcommand{\RQfourModelList}{SmolLM2-135M, SmolLM2-360M and Qwen2.5-0.5B}

\newcommand{\RQoneGTotalCI}{\textminus 1.61\,\ci{\textminus 2.12}{\textminus 1.14}}

\newcommand{\RQoneGAbstractCI}{\textminus 0.39\,\ci{\textminus 0.54}{\textminus 0.24}}

\newcommand{\RQoneAbstractRemoved}{76\%}

\newcommand{\RQoneAbstractLowersN}{5}
\newcommand{\RQoneAbstractLowersList}{SmolLM2-135M, SmolLM2-360M, Qwen3-0.6B, SmolLM2-1.7B and Qwen3-1.7B}
\newcommand{\RQoneAbstractRaisesList}{Qwen2.5-0.5B and Llama-3.2-1B}

\newcommand{\PriorSlopeFail}{0.18}
\newcommand{\PriorSlopeFailCI}{+0.18\,\ci{+0.14}{+0.21}}

\newcommand{\PriorSlopeNeutCI}{+0.24\,\ci{+0.19}{+0.28}}

\newcommand{\PriorSlopeAbstractCI}{+0.75\,\ci{+0.72}{+0.78}}

\newcommand{\PriorSlopeN}{2079}
\newcommand{\PriorSlopeMaxModel}{0.35}
\newcommand{\PriorSlopeMinModel}{0.02}

\newcommand{\PriorSlopeFailBaseCI}{+0.45\,\ci{+0.37}{+0.55}}

\newcommand{\PriorSlopeFailRagenCI}{+0.09\,\ci{+0.04}{+0.13}}

\newcommand{\PriorBinLowGCI}{\textminus 2.62\,\ci{\textminus 3.08}{\textminus 2.16}}
\newcommand{\PriorBinLowN}{652}
\newcommand{\PriorBinLowGreedyPre}{0\%}
\newcommand{\PriorBinLowGreedyPost}{22\%}
\newcommand{\PriorBinLowGreedyAbstract}{0\%}

\newcommand{\PriorBinHighGCI}{+1.68\,\ci{+1.46}{+1.91}}
\newcommand{\PriorBinHighN}{469}
\newcommand{\PriorBinHighGreedyPre}{91\%}
\newcommand{\PriorBinHighGreedyPost}{38\%}
\newcommand{\PriorBinHighGreedyAbstract}{60\%}
\newcommand{\PriorModelLowShare}{2\%}
\newcommand{\PriorModelHighShare}{44\%}

\newcommand{\PriorModelGreedyPre}{66\%}
\newcommand{\PriorModelGreedyPost}{32\%}
\newcommand{\PriorModelGreedyAbstract}{36\%}
\newcommand{\PriorSharedLowShare}{48\%}
\newcommand{\PriorSharedHighShare}{10\%}

\newcommand{\AgentEpisodes}{8}
\newcommand{\AgentModels}{SmolLM2-360M and Qwen2.5-0.5B}
\newcommand{\AgentNumCells}{6}
\newcommand{\AgentVerbatimRepeat}{72\%}
\newcommand{\AgentVerbatimIllegal}{81\%}

\newcommand{\AgentInstrRepeat}{71\%}

\newcommand{\AgentDropRepeat}{94\%}

\newcommand{\AgentAbstractRepeat}{75\%}

\newcommand{\AgentVerbatimBanRepeat}{3\%}
\newcommand{\AgentVerbatimBanIllegal}{79\%}

\newcommand{\AgentAbstractBanRepeat}{0\%}

\newcommand{\AgentBanRepeatMax}{12\%}
\newcommand{\AgentCellsWithWins}{2}

\newcommand{\AgentStepLimitShare}{68\%}
\newcommand{\AgentIllegalOverall}{81\%}

\title{Feedback That Backfires in Games:\\
Do LLM Players Repeat the Move They Just Watched Fail?}

\author{%
  Esmail Gumaan\\
  Faculty of Computer Science and Mathematics\\
  University of Passau, Germany\\
  \texttt{gumaan01@ads.uni-passau.de}
}

\begin{document}

\maketitle

\begin{abstract}
When an LLM agent's move fails, the harness writes the move and the failure
message into the transcript and asks again. In tool calling, this record makes
small models more likely to repeat the failed action. We test whether the same
holds in games (Sokoban, Connect Four, Othello), where a move can fail by
being illegal or by being legal but losing, both verified exactly. On fixed
probe items built by perturbing a correct move, the record backfires: pooled
over \NumModelsRQone\ instruction-tuned models (\SmallestModelRQone--\LargestModelRQone)
it makes the failed move more likely (corrective gain $\Gain = \RQoneG$ nats
per token), through the move's surface form, for illegal moves but not for
losing ones, and not for every model. An exploratory analysis shows what sets the sign:
after the record, the failed move's log-probability barely depends on how
likely it was before (slope \PriorSlopeFail). Moves the model would not have
played become more likely, and greedy decoding starts to produce them; moves
it was about to play become less likely, but greedy decoding still repeats
\PriorBinHighGreedyPost\ of them. A description of the failure that omits the
move string keeps the model's preference instead. Along the OLMo 2 1B
post-training ladder the inversion is present in the base model and largely
removed at the DPO stage. RAGEN's default models invert in RAGEN's own Sokoban
format before any RL. A ``do not repeat'' instruction does nothing, a clean
retry is the worst harness, and a decoder ban stops exact repeats but not
illegal play. All measurements ran in float32 on a free Colab GPU, which first
reproduced the original paper's number.
\end{abstract}

\section{Introduction}
\label{sec:intro}

When an LLM plays a game through a harness, some of its moves fail. The
harness then does what agent harnesses do with a failed tool call: it writes
the move into the transcript, adds the referee's message (\mv{Illegal move:
that column is already full.}) and asks for another move. The message is
ground truth, and a player that reads it should not try the same move again.
Yet LLM players are known to get stuck repeating rejected moves; in
XiangqiBench, rejected moves make up a third of all calls in failed trials of
one protocol \citep{chai2026xiangqibench}.

In earlier work we measured what such a failure record does to a small
model's next action in tool calling and program repair
\citep{gumaan2026backfire}. Defining the \emph{corrective gain} $\Gain$ of the
record as the drop in log-probability of re-emitting the failed action, we
found it negative for every instruction-tuned model we tested: the record made
the failed call \emph{more} likely, mostly because the call's surface form was
now in the context. This paper asks whether that holds for game play, where
the setting allows three things tool calling did not.

First, a move can fail in two ways: it can be \emph{illegal}, or legal but
\emph{losing}, and both can be verified exactly (a one-ply check in Connect
Four, an exhaustive search for Sokoban deadlocks). Second, the same move can be
written in different notations, which varies the surface form while holding
the decision fixed. Third, a completion format lets us probe base models, so
the measurement can follow one model through the stages of post-training. We
use Sokoban, Connect Four and Othello, and ask four questions:
\textbf{RQ1}, does the failure record backfire in games, for illegal and for
losing moves? \textbf{RQ2}, which stage of training produces it, along the
public OLMo 2 1B ladder from base to Instruct \citep{olmo2025olmo2}?
\textbf{RQ3}, how do the default models of RAGEN \citep{wang2025ragen}, the
standard multi-turn RL framework for these games, behave in RAGEN's own prompt
format? \textbf{RQ4}, do the original paper's remedies transfer?

\paragraph{Findings.}
\begin{itemize}[leftmargin=*,itemsep=2pt,topsep=2pt]
  \item On fixed probe items built by perturbing a correct move, as in the
  original, the record backfires: pooled over \NumModelsRQone\ instruction-tuned
  models it makes the failed move more likely, $\Gain = \RQoneG$ nats per token
  (Section~\ref{sec:rq1}), carried by the surface form ($\copyterm =
  \RQoneCopy$) rather than the failure message ($\semterm = \RQoneSem$). The
  effect holds for illegal moves and not
  for losing ones, and it is not universal: \RQoneNumNeg\ of the
  \NumModelsRQone\ models invert clearly, and \RQonePosList\ does the opposite.
  \item The sign depends on how plausible the failed move was
  (Section~\ref{sec:prior}). After the record, the failed move's
  log-probability barely depends on its value before the attempt (slope
  \PriorSlopeFail\ per nat). Moves the model would not have played become more
  likely; greedy decoding, which never produced them, produces
  \PriorBinLowGreedyPost\ of them after the record. Moves the model was likely
  to play become less likely, but greedy decoding still repeats
  \PriorBinHighGreedyPost\ of them. This flattening comes with writing the
  move: a description of the failure without the move string keeps the
  model's preference.
  \item Along the OLMo 2 ladder the inversion is already present in the base
  model, and the DPO stage removes most of it (Section~\ref{sec:rq2}).
  \item In RAGEN's format, both untrained default models invert on illegal
  Sokoban actions, with RAGEN's own feedback as well (Section~\ref{sec:rq3}).
  \item A ``do not repeat'' instruction does nothing and a clean retry is the
  worst harness for repetition, as in tool calling. The runtime-generated
  description removes most of the inversion on average but not for every
  model; a decoder ban removes exact repeats but not illegal play
  (Section~\ref{sec:rq4}).
\end{itemize}

We could not run the RL half of RQ3 or a 7B scale check; everything reported
here ran in float32 on a free Colab GPU after the GPU reproduced the original
paper's headline number (Section~\ref{sec:setup}). Code, items, raw
per-item records and run metadata are available at \RepoURL.

\section{Related work}
\label{sec:related}

\paragraph{The measurement this paper extends.} \citet{gumaan2026backfire}
define the corrective gain and find it negative for six instruction-tuned
models (135M--1.7B) in simulated tool calling and MBPP program repair, with
most of the damage carried by the failed call's surface form. Replacing the
call with a runtime-generated description of the failure removed most of the
inversion; an explicit ``do not repeat'' instruction and a clean retry did
not. That study left three things open, and they define this one: whether the
effect appears outside tool use, in a setting where a failure can be
\emph{illegal} or merely \emph{bad}; which stage of training produces it; and
whether training on the task, rather than editing the harness, removes it.

\paragraph{LLMs that repeat themselves in decisions.}
\citet{schmied2025greedy} identify greediness, frequency bias and a
knowing-doing gap in LLM agents on bandits and tic-tac-toe, and show that RL
fine-tuning on self-generated rationales reduces but does not remove them.
Their frequency bias is a tendency to repeat the action that is most frequent
in the context; they do not condition on failure feedback, and they measure
behaviour, not log-probabilities. We ask the narrower counterfactual question
of what a \emph{failure record} does to the probability of the failed move,
which their setup does not separate from frequency.

\paragraph{Multi-turn RL for LLM agents.} RAGEN and its StarPO objective
\citep{wang2025ragen} train small models on Sokoban, FrozenLake and bandits
and describe an ``Echo Trap'', a training-time collapse into repetitive
reasoning with falling reward variance. That is a property of the training
dynamics; ours is an inference-time property of the transcript. We probe
RAGEN's default models in RAGEN's own prompt format, which is where RL on the
game would start; measuring the same models after training is left open.

\paragraph{Games as agent benchmarks.} XiangqiBench \citep{chai2026xiangqibench}
evaluates frontier LLMs as closed-loop players; in its restricted-observation
protocol, rejected moves make up a third of all calls in failed trials and
about a fifteenth in successful ones. It reports the symptom without
isolating a cause, and we use it as motivation only, since our experiments
include no chess-family game. BALROG \citep{paglieri2024balrog}, TextArena
\citep{guertler2025textarena}, SmartPlay \citep{wu2023smartplay} and GTBench
\citep{duan2024gtbench} score policies across many games; like the agent
benchmarks the original discussed, they are built to rank models, not to
expose the counterfactual observations a decomposition needs.

\paragraph{Feedback in the transcript.} ReAct \citep{yao2023react},
Reflexion \citep{shinn2023reflexion} and Self-Refine
\citep{madaan2023selfrefine} append actions, observations or critiques to a
growing transcript on the premise that feedback is informative. Work on
context contamination \citep{yang2026contamination} finds that retries in a
context that still holds the failed attempt fail more often and prescribes
clearing the context; the original showed that this clean retry was the worst
harness under greedy decoding.

\paragraph{Copying.} Induction heads implement prefix matching and copying
\citep{olsson2022induction}, and repetition is self-reinforcing in open-ended
generation \citep{xu2022breaktheloop}. If the inversion is copying, it should
already be present in a base model, before any instruction tuning; the
post-training ladder of OLMo 2 \citep{olmo2025olmo2}, whose checkpoints differ
only in post-training, makes that testable.

\paragraph{Negative instructions.} Naming something raises its probability
even under an instruction to avoid it
\citep{castricato2024pinkelephants,negation2025pinkelephant}; the original's
``do not repeat'' result is an instance, and we test whether it transfers.

\section{Measuring the gain in games}
\label{sec:method}

\subsection{The quantity}

We use the original paper's definitions unchanged \citep{gumaan2026backfire}.
Fix a decision point: a context $\ctxpre$ that ends with a position $s$ and a
request for a move, a move $\afail$ that fails at $s$, the runtime's
observation $o^{\times}$, and a correct move $\agold$. The \emph{corrective
gain} of the failure record is
\begin{equation}
\Gain(\afail) = \log \pi(\afail \mid \ctxpre) - \log \pi\big(\afail \mid \ctxpre \oplus (\afail, o^{\times})\big),
\label{eq:gain}
\end{equation}
where $\log\pi(a \mid c)$ is the summed token log-probability of the move
string and $\oplus$ appends one agent turn and one runtime turn to the
transcript. $\Gain > 0$ is what a harness is for: after the failure is
recorded the failed move is less likely. $\Gain < 0$ is \emph{feedback
inversion}. With a valence-free observation $o^{\varnothing}$
(\mv{Move received.}) the gain splits into a surface-form and a semantic term,
\begin{align}
\copyterm &= \log \pi\big(\afail \mid \ctxpre \oplus (\afail, o^{\varnothing})\big) - \log \pi(\afail \mid \ctxpre),\\
\semterm &= \log \pi\big(\afail \mid \ctxpre \oplus (\afail, o^{\times})\big) - \log \pi\big(\afail \mid \ctxpre \oplus (\afail, o^{\varnothing})\big),
\end{align}
with $-\Gain = \copyterm + \semterm$ by construction. $\copyterm$ is what
merely having written the move does; $\semterm$ is what saying that it failed
adds. A counterfactual success observation $o^{\checkmark}$ (\mv{Move
accepted.}) gives the polarity contrast
$\Delta_{\mathrm{pol}} = \log\pi(\afail \mid \ctxpre\oplus(\afail,o^{\times})) - \log\pi(\afail \mid \ctxpre\oplus(\afail,o^{\checkmark}))$.
We report every quantity in nats and, because move strings differ in length
across notations, per token of the move (the original's normalisation), and we
add the original's two readouts: the probability of the failed move
normalised over a fixed four-move candidate set $\{\afail, \agold, d_1, d_2\}$,
and whether greedy decoding would reproduce $\afail$ token for token, which is
read off the argmax path of the teacher-forced pass at no extra cost.

\subsection{Two kinds of failure}

Games let us ask a question tool calling could not: does it matter
\emph{why} the move failed?

\paragraph{Illegal moves.} The runtime rejects the move and the position is
unchanged: a full or non-existent column in Connect Four; an occupied square,
a placement that flips nothing, or an off-board square in Othello; an unknown
action, a walk into a wall, or a push against a wall or box in Sokoban.

\paragraph{Legal but losing moves.} The move is accepted and is a mistake,
and the harness lets the agent retry from the restored position. The loss is
verified exactly, never judged: in Connect Four a move is \emph{losing} if,
after it, the opponent has an immediate winning reply (checked by trying every
reply; a move that wins at once is never labelled losing); in Sokoban a push
is a \emph{deadlock} if the room was solvable before it and is not after it
(an exact breadth-first search over player and box positions). The message
says so and that the position has been restored.

\subsection{Conditions}
\label{sec:method:conditions}

All conditions of an item share the system prompt, the goal, the successful
prefix and the position $s$, byte for byte. Every post-attempt condition
re-shows the same position $s$ after its message, so conditions differ in
exactly one thing, the record of the attempt (Table~\ref{tab:conditions} in
the appendix lists them all).
The core set is the original's: \texttt{pre} (no attempt; also the
clean-retry context), \texttt{fail} (the move and its failure message),
\texttt{succ}, \texttt{neut}, and \texttt{abstract} (a runtime-generated
description of the failure, with no move string). The remedy and control set
adds \texttt{fail\_instr} (a ``do not repeat'' instruction in the system
prompt), \texttt{fail\_echo} (the message also quotes the move),
\texttt{fail\_k2} (the move fails twice), \texttt{fail\_other} (a different
failing move failed; the scored move was never written) and
\texttt{abstract\_min} (a bare failure marker). In the RAGEN format a further
condition, \texttt{fail\_native}, uses RAGEN's own feedback: its
invalid-action warning, or nothing at all after a blocked move.

Two rules from the original keep the decomposition clean. First, the
\emph{primary} failure message never contains the move string
(\mv{Illegal move: that column is already full.}, not
\mv{column 4 is full}); otherwise \texttt{fail} would hold the move twice and
\texttt{neut} once, and occurrence count would leak into $\semterm$. Naming the
move is the separate \texttt{fail\_echo} manipulation. Second, the abstract
description contains no move string and no digit in any notation: its counter
is an ordinal word (\mv{[first attempt failed: the chosen column was already
full; that move is not repeatable]}), because a digit would \emph{be} the move
in the numeric notations.

\subsection{Which failed move to probe}
\label{sec:method:sources}

The original fixed the failing action by a perturbation operator so that every
model was probed on byte-identical items; sampling each model's own mistakes
would confound model with item difficulty. We keep that as the primary design
and add two other sources, recording the source of every item.
(b) \emph{Perturbation} items apply an operator to the correct move: an
adjacent or mirrored column; an adjacent, transposed or mirrored square; the
opposite or a perpendicular direction, or a compass synonym
(\mv{North} for \mv{Up}). An item is kept only if the result fails at $s$, and
items are balanced over failure types.
(c) \emph{Random} items draw the failing move uniformly from every failing
move expressible at $s$, a control.
(a) \emph{Model-proposed} items are the realistic case and are model-specific
by construction: a proposal pass scores every move of the game, legal and
illegal, at \texttt{pre}, and the item is the highest-ranked failing move among
that model's top three. They are analysed as a robustness check, never pooled
with the shared items in a cross-model comparison.

\subsection{Mechanism test: notation}
\label{sec:method:notation}

If the inversion is carried by the move's surface form, the way a move is
written should change its size. Each game has three notations: a short code
(\mv{4}, \mv{d3}, \mv{U}), a verbose phrase (\mv{drop a disc into column 4},
\mv{place a disc on column d, row 3}, \mv{move the player one square up}) and a
plain zero-based index (\mv{3}, \mv{19}, \mv{0}). Othello indices are
zero-padded so that no move string is a prefix of another, which the
unterminated scoring of the original requires. The board, rules and messages
are otherwise identical across notations; only Connect Four's column labels
follow the notation. For every item and model we record the move's length in
tokens and its token overlap with the legal alternatives at $s$ (the largest
multiset overlap with any legal move), and regress the per-token gain on
notation, length and overlap with random effects for position and model.

\subsection{Formats, and why there are three}
\label{sec:method:formats}

A post-training ladder starts at a base model, which has no chat template. We
therefore probe every checkpoint of the ladder in one \emph{completion}
format: the rules, a blank line, then \mv{Referee:} and \mv{Player:} lines,
ending in \mv{Player:}, with the scored continuation \mv{\textvisiblespace
move}, exactly as the move appears once written. Chat checkpoints are also
probed in their own \emph{chat} template, as a robustness check and for the
cross-model study. RAGEN's default models are also probed in the \emph{RAGEN}
format: RAGEN's system prompt, \mv{Turn k:} and \mv{State:} blocks,
\mv{Reward:} lines, answers in \mv{<answer>} tags, and our messages in RAGEN's
warning slot. Our tests extract RAGEN's own prompt-building functions from its
source and check that the messages we build are identical, so the models are
probed on exactly the prompt shape that RL on the game would train them on. All
three formats are rendered from one transcript representation, so a condition
is defined once and cannot drift between formats.

\subsection{Scoring and statistics}
\label{sec:method:scoring}

Scoring is the original's code, vendored unchanged: teacher-forced, in
float32, with a prefix key/value cache that crops to the longest common prefix
of consecutive contexts and a finite-logits check at load and on every forward
pass. On SmolLM2-135M a sample of items in each of the three formats is
re-scored without the cache and must agree to $10^{-3}$ nats. Items that share a position share a context, so every interval
is a 95\% cluster bootstrap over positions ($10{,}000$ resamples), and
contrasts between conditions, checkpoints or harnesses are paired item by item.
The bootstrap is the original's estimator; we compute it without its
per-resample loop, from the same random draws, and a test checks that both give
the same intervals.

\paragraph{The new code reproduces the old number.} Before any game was
scored, the original paper's 100 tool-calling items were run through the new
pipeline on SmolLM2-135M, first on the original's laptop CPU and then on the
GPU used for every result here. The CPU run returns $\Gain = \RegressionG$
nats, against \PaperGOriginal\ in the original, with every stored
log-probability reproduced to within \RegressionMaxDiff\ nats; the GPU run
returns \RegressionGPUG, within \RegressionGPUMaxDiff\ nats of the CPU rows.

\section{Experimental setup}
\label{sec:setup}

\subsection{Games}

We use three games and no others; none involves chance or a chess-family
piece set.

\paragraph{Sokoban.} Rooms come from RAGEN's own generator
\citep{wang2025ragen}, vendored unchanged, with its \emph{SimpleSokoban}
parameters: $6\times6$, one box, text grid. Step semantics and rewards follow
gym-sokoban \citep{schrader2018gymsokoban}, which RAGEN wraps. The agent is
shown the moves it has made so far (at most \SokobanMaxPrefix), so a probe
context starts at the room's first turn, as a RAGEN transcript does. Failure
types: unknown action, walk into a wall, push against a wall or box (illegal),
and deadlock push (losing).

\paragraph{Connect Four} ($6\times7$). Positions are taken from seeded
self-play between random and one-ply-greedy players. The context shows the
position two agent turns earlier and the two most recent agent moves with the
opponent's replies. Failure types: full column and non-existent column
(illegal), and losing move (legal; the opponent can win at once).

\paragraph{Othello} ($8\times8$). Positions from seeded self-play as above,
with the same two-turn window; windows that contain a pass are skipped so that
turns alternate. Failure types: occupied square, placement that flips nothing,
off-board square (all illegal). We do not label losing moves in Othello.

\subsection{Items}

Each game contributes one decision point per position:
\NumPointsSokoban\ for Sokoban, \NumPointsConnectFour\ for Connect Four and
\NumPointsOthello\ for Othello. Perturbation items are balanced over failure
types (\PerTypeQuota\ per type) and \NRandomItems\ random-source items are
added per game; Appendix~\ref{app:items} gives the counts by failure type,
source and operator. A run scores a prefix of one fixed order of the
positions that cycles over failure types: 40 positions per game for the RQ1
models up to 0.6B and for the remedy study, 30 for the four larger RQ1
models, 20 for each OLMo 2 checkpoint, and 60 Sokoban positions in the RAGEN
format. These sample sizes were set for a laptop CPU
(Appendix~\ref{app:repro}). A smaller sample is a prefix of a larger one, so
a model scored on fewer positions sees a subset of the same items. Every
decision point is rendered in every notation, so
the notation contrast is within position. All messages are fixed in a
versioned file (Appendix~\ref{app:messages}). A check found that the failed
move is not always new to the context: in \FailedInPrefixConnectFour\ of the
Connect Four and \FailedInPrefixSokoban\ of the Sokoban decision points the
failed move equals one of the agent's own earlier moves (in Sokoban, the
deadlock push often continues the previous direction). Tool calls in the
original never had this property; we record it per item and use it as a
covariate.

\subsection{Models}

\paragraph{Instruction-tuned models (RQ1).} The original paper's models for
continuity, plus two: \ModelListRQone, each probed in its own chat template
\citep{allal2025smollm2,qwen2024qwen25,qwen2025qwen3,grattafiori2024llama3}.

\paragraph{The OLMo 2 1B post-training ladder (RQ2).} The five public
checkpoints of \texttt{allenai/OLMo-2-0425-1B}: base, SFT, DPO, RLVR and the
released Instruct model \citep{olmo2025olmo2}, which follow the Tulu~3 recipe
\citep{lambert2024tulu3}. All five share one tokenizer and one architecture
and differ only in post-training, which is what makes the stage comparison
interpretable. Every checkpoint is probed in the completion format, the four
post-trained ones also in their chat template.

\paragraph{The RAGEN format (RQ3).} \texttt{Qwen2.5-0.5B-Instruct} and
\texttt{Qwen2.5-1.5B-Instruct}, RAGEN's default models and members of the
RQ1 set, probed on Sokoban in RAGEN's own SimpleSokoban prompt, in no-think
mode with one action per turn, the setting in which RAGEN trains them
\citep{wang2025ragen}. These are the untrained starting points of RL on the
game. Training them with StarPO and probing again, which our proposal planned
as the second half of RQ3, needed more GPU time than we had
(Section~\ref{sec:limitations}).

Every checkpoint is pinned to a Hugging Face revision (Table~\ref{tab:models}
in the appendix).

\subsection{Hardware and numerics}

All reported scoring and free play ran on one \GPUName\ in Google Colab's free
tier, in float32 with TF32 matrix multiplication \TFThirtyTwo, under PyTorch
\TorchVersion\ and Transformers \TransformersVersion, the Transformers version
of the original run of record. Items were built, and the pipeline was
developed and first validated, on the original paper's laptop (\CPUName,
PyTorch \CPUTorchVersion). Before any game was scored, the GPU had to
reproduce the original paper's number (Section~\ref{sec:method:scoring};
tolerance \RegressionGPUTol\ nats per log-probability) and cached and uncached
scoring had to agree on it in all three formats. A small CPU pilot run during
development is not used for any reported number, so every result in this
paper comes from one device.

\section{RQ1: does the failure record backfire in games?}
\label{sec:rq1}

\begin{table}[t]
\centering
\scriptsize
\setlength{\tabcolsep}{2.5pt}
\resizebox{\linewidth}{!}{%
\begin{tabular}{l c c c c c c r}
\toprule
& \multicolumn{3}{c}{$G$ per move token, by game} & \multicolumn{3}{c}{pooled over games} & \\
\cmidrule(lr){2-4}\cmidrule(lr){5-7}
Model & Sokoban & Connect Four & Othello & $G$/tok & $\Delta_{\mathrm{copy}}$/tok & $\Delta_{\mathrm{sem}}$/tok & $G<0$ \\
\midrule
SmolLM2-135M & \textbf{\textminus 2.35\,\ci{\textminus 2.84}{\textminus 1.86}} & \textbf{\textminus 1.34\,\ci{\textminus 1.98}{\textminus 0.71}} & \textbf{\textminus 1.22\,\ci{\textminus 1.47}{\textminus 0.98}} & \textbf{\textminus 1.64\,\ci{\textminus 1.93}{\textminus 1.35}} & \textbf{1.47\,\ci{1.16}{1.78}} & \textbf{0.17\,\ci{0.09}{0.24}} & 92\% \\
SmolLM2-360M & \textbf{\textminus 1.14\,\ci{\textminus 1.74}{\textminus 0.58}} & \textbf{\textminus 0.95\,\ci{\textminus 1.42}{\textminus 0.50}} & \textbf{\textminus 1.18\,\ci{\textminus 1.41}{\textminus 0.98}} & \textbf{\textminus 1.09\,\ci{\textminus 1.36}{\textminus 0.85}} & \textbf{1.11\,\ci{0.86}{1.37}} & \textminus 0.02\,\ci{\textminus 0.10}{0.07} & 74\% \\
Qwen2.5-0.5B & \textminus 0.33\,\ci{\textminus 1.54}{0.85} & \textminus 0.20\,\ci{\textminus 1.05}{0.64} & 0.39\,\ci{\textminus 0.01}{0.76} & \textminus 0.05\,\ci{\textminus 0.58}{0.44} & \textminus 0.33\,\ci{\textminus 0.86}{0.23} & \textbf{0.38\,\ci{0.13}{0.65}} & 40\% \\
Qwen3-0.6B & \textbf{\textminus 5.24\,\ci{\textminus 6.51}{\textminus 3.97}} & \textbf{\textminus 6.68\,\ci{\textminus 7.75}{\textminus 5.52}} & \textbf{\textminus 4.45\,\ci{\textminus 4.95}{\textminus 3.96}} & \textbf{\textminus 5.45\,\ci{\textminus 6.08}{\textminus 4.85}} & \textbf{5.62\,\ci{5.02}{6.25}} & \textbf{\textminus 0.17\,\ci{\textminus 0.21}{\textminus 0.13}} & 92\% \\
Llama-3.2-1B & 0.48\,\ci{\textminus 0.18}{1.06} & 0.36\,\ci{\textminus 0.61}{1.28} & \textbf{1.15\,\ci{0.79}{1.49}} & \textbf{0.66\,\ci{0.26}{1.05}} & \textbf{\textminus 0.52\,\ci{\textminus 0.95}{\textminus 0.08}} & \textbf{\textminus 0.15\,\ci{\textminus 0.26}{\textminus 0.02}} & 23\% \\
Qwen2.5-1.5B & \textbf{\textminus 2.02\,\ci{\textminus 3.73}{\textminus 0.46}} & \textminus 1.72\,\ci{\textminus 3.94}{0.36} & \textbf{1.12\,\ci{0.53}{1.70}} & \textminus 0.87\,\ci{\textminus 1.85}{0.05} & 0.72\,\ci{\textminus 0.11}{1.59} & 0.15\,\ci{\textminus 0.13}{0.42} & 44\% \\
SmolLM2-1.7B & \textminus 0.74\,\ci{\textminus 1.65}{0.11} & \textminus 0.44\,\ci{\textminus 1.14}{0.22} & \textbf{\textminus 0.73\,\ci{\textminus 0.99}{\textminus 0.47}} & \textbf{\textminus 0.64\,\ci{\textminus 1.02}{\textminus 0.26}} & \textminus 0.19\,\ci{\textminus 0.55}{0.20} & \textbf{0.82\,\ci{0.63}{1.02}} & 58\% \\
Qwen3-1.7B & \textminus 0.11\,\ci{\textminus 2.95}{2.65} & \textbf{\textminus 3.57\,\ci{\textminus 5.27}{\textminus 1.91}} & 0.51\,\ci{\textminus 1.12}{2.05} & \textminus 1.06\,\ci{\textminus 2.32}{0.22} & \textbf{2.04\,\ci{1.12}{2.98}} & \textbf{\textminus 0.98\,\ci{\textminus 1.76}{\textminus 0.21}} & 50\% \\
\bottomrule
\end{tabular}}
\caption{RQ1: corrective gain of a failure record, instruction-tuned models in their chat template, short notation, shared items (perturbation and random sources). Nats per move token; $G<0$ is feedback inversion. Cells: mean and 95\% cluster-bootstrap interval over positions; bold intervals exclude zero. Last column: share of items with $G<0$.}
\label{tab:rq1}
\end{table}

Table~\ref{tab:rq1} gives the corrective gain of a failure record for each
instruction-tuned model, per game and pooled over the three games, on the
shared items. Pooled over the \NumModelsRQone\ models, the record
\emph{raises} the probability of the move that just failed:
$\Gain = \RQoneGCI$ nats per move token, negative on \RQoneFracNeg\ of the
\RQoneNItems\ (model, item) pairs. On a probability scale, the normalised
probability of the failed move over the four-move candidate set rises from
\RQonePRepeatPre\ before the attempt to \RQonePRepeatPost\ after it, and
greedy decoding reproduces the failed move token for token on
\RQoneGreedyPost\ of items after the failure, against \RQoneGreedyPre\ before
it.

\begin{figure}[t]
  \centering
  \includegraphics[width=0.48\linewidth]{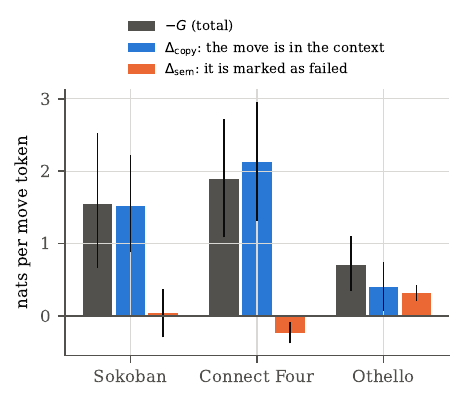}
  \caption{The decomposition $-\Gain = \copyterm + \semterm$ by game, pooled
  over the RQ1 models (shared items, nats per move token, 95\% intervals
  clustered on position). Positive bars push towards repeating the failed
  move.}
  \label{fig:decomposition}
\end{figure}

As in tool calling, the increase is carried by the surface form. Writing the
move into the transcript with a valence-free acknowledgement accounts for
$\copyterm = \RQoneCopyCI$ nats per token; replacing that acknowledgement with
the failure message changes the failed move's log-probability by a further
$\semterm = \RQoneSemCI$, an interval that includes zero
(Figure~\ref{fig:decomposition}).

\paragraph{Not every model.} The pooled number hides a spread. \RQoneNumNeg\
of the \NumModelsRQone\ models invert with intervals entirely below zero
(\RQoneNegList), and \RQoneNumNull\ more have point estimates below zero but
intervals that include it (\RQoneNullList). \RQonePosList\ goes the other way:
its gain is positive, \RQoneGLlamaCI, and its copy term is negative,
\RQoneCopyLlamaCI. For this model, seeing its own move in the transcript makes
that move \emph{less} likely. The pooled mean is also strongly influenced by
Qwen3-0.6B, whose inversion (\RQoneGQwenThreeSCI) is the largest we measured.
In the original's tool-calling study every model inverted; in games,
\RQoneNumNegPoint\ of the \NumModelsRQone\ point estimates are below zero.

\begin{figure}[t]
  \centering
  \includegraphics[width=\linewidth]{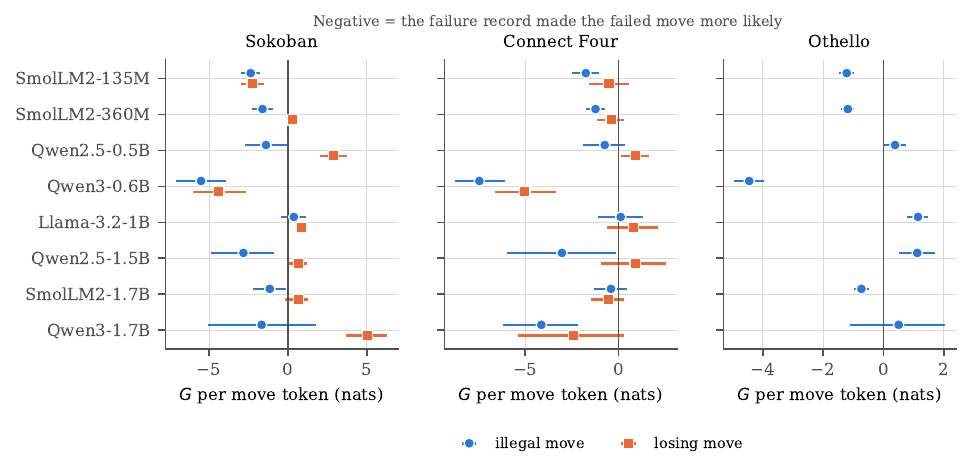}
  \caption{RQ1 by model, game and failure class (chat template, short
  notation, shared items). Points: mean $\Gain$ per move token; bars: 95\%
  cluster-bootstrap intervals over positions. Othello has no losing items.}
  \label{fig:rq1}
\end{figure}

\paragraph{Illegal versus losing moves.} Figure~\ref{fig:rq1} and
Table~\ref{tab:rq1-class} separate the two kinds of failure. Illegal moves
invert clearly: $\Gain = \RQoneIllegalGCI$ nats per token, negative on
\RQoneIllegalFracNeg\ of items, with a copy term of \RQoneIllegalCopyCI.
Legal but losing moves invert much less, $\Gain = \RQoneLosingGCI$
(\RQoneLosingNPositions\ positions), an interval that includes zero. Their
copy term is still positive (\RQoneLosingCopyCI), but greedy decoding repeats
a losing move less often after the failure (\RQoneLosingGreedyPost) than
before it (\RQoneLosingGreedyPre). By game, losing moves in Connect Four give
\RQoneLosingGConnectFourCI\ and deadlock pushes in Sokoban
\RQoneLosingGSokobanCI.

\begin{table}[t]
\centering
\scriptsize
\setlength{\tabcolsep}{3pt}
\begin{tabular}{l l c c c c c r}
\toprule
Game & Failure & $G$/tok & $\Delta_{\mathrm{copy}}$/tok & $\Delta_{\mathrm{sem}}$/tok & $P(\text{repeat})$ & greedy repeat & $n$ \\
\midrule
Sokoban & illegal & \textbf{\textminus 2.12\,\ci{\textminus 3.28}{\textminus 1.04}} & \textbf{1.89\,\ci{1.11}{2.76}} & 0.23\,\ci{\textminus 0.17}{0.62} & 0.25 $\to$ 0.39 & 18\% $\to$ 35\% & 212 \\
Sokoban & losing & 0.23\,\ci{\textminus 0.26}{0.60} & 0.35\,\ci{\textminus 0.11}{0.79} & \textbf{\textminus 0.58\,\ci{\textminus 0.87}{\textminus 0.18}} & 0.45 $\to$ 0.38 & 38\% $\to$ 19\% & 68 \\
Connect Four & illegal & \textbf{\textminus 2.40\,\ci{\textminus 3.53}{\textminus 1.33}} & \textbf{2.78\,\ci{1.78}{3.87}} & \textbf{\textminus 0.39\,\ci{\textminus 0.51}{\textminus 0.26}} & 0.20 $\to$ 0.30 & 7\% $\to$ 20\% & 188 \\
Connect Four & losing & \textbf{\textminus 0.84\,\ci{\textminus 1.48}{\textminus 0.22}} & \textbf{0.78\,\ci{0.01}{1.53}} & 0.06\,\ci{\textminus 0.25}{0.35} & 0.26 $\to$ 0.33 & 16\% $\to$ 22\% & 92 \\
Othello & illegal & \textbf{\textminus 0.70\,\ci{\textminus 1.10}{\textminus 0.34}} & \textbf{0.39\,\ci{0.07}{0.75}} & \textbf{0.31\,\ci{0.20}{0.42}} & 0.21 $\to$ 0.34 & 1\% $\to$ 18\% & 280 \\
\bottomrule
\end{tabular}
\caption{RQ1 by game and failure class, pooled over the instruction-tuned models (intervals clustered on position). $P(\text{repeat})$: probability of the failed move normalised over the four-move candidate set, before $\to$ after the failure record. Greedy repeat: share of items on which greedy decoding reproduces the failed move token for token, before $\to$ after.}
\label{tab:rq1-class}
\end{table}

\paragraph{The model's own mistakes go the other way.} On the realistic items,
where the failed move is the highest-ranked failing move among the model's own
top three, the sign flips: $\Gain = \RQoneModelSrcGCI$ nats per token
(\RQoneModelSrcNItems\ items), negative on only \RQoneModelSrcFracNeg. These
items differ between models by construction and are never pooled with the
shared ones, but the difference is too large to be a detail.
Section~\ref{sec:prior} shows what separates the two kinds of item: how
plausible the failed move was before the attempt.

\paragraph{Answer.} H1 holds on the fixed items: a recorded illegal move makes
that move more likely, through its surface form, for most but not all of the
models. H2 does not hold: for legal but losing moves the pooled gain is
indistinguishable from zero. On the models' own likely mistakes the record
lowers the failed move's probability instead.

\section{What sets the sign: the record flattens the model's preference}
\label{sec:prior}

This analysis was not part of our plan. We report it because it explains the
opposite signs of Section~\ref{sec:rq1}, and because it changes what the
inversion means for an agent.

\begin{figure}[t]
  \centering
  \includegraphics[width=\linewidth]{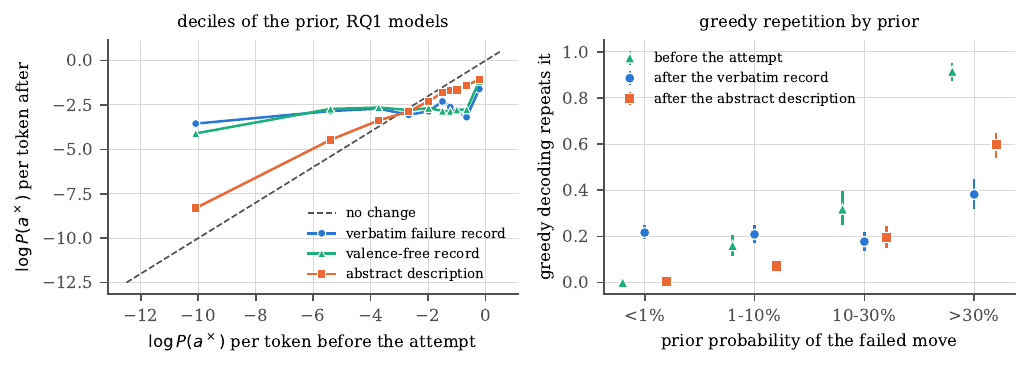}
  \caption{Exploratory. Left: mean per-token log-probability of the failed
  move after each kind of record, against its value before the attempt
  (deciles of the latter; RQ1 models, chat template, all item sources). The
  dashed line is no change. Right: share of items on which greedy decoding
  reproduces the failed move, by its prior probability; bars are 95\% cluster
  intervals over positions.}
  \label{fig:prior}
\end{figure}

Call a failed move's log-probability before the attempt, $\log\pi(\afail
\mid \ctxpre)$, its \emph{prior}. If a failure record shifted every failed move
by the same amount, the log-probability after the record would follow the
prior with slope one. It does not (Figure~\ref{fig:prior}, left). Over all
\PriorSlopeN\ RQ1 items, shared and model-proposed, the per-token
log-probability of the failed move after the verbatim failure record has a
slope of \PriorSlopeFailCI\ on its prior; per model the slope lies between
\PriorSlopeMinModel\ and \PriorSlopeMaxModel. After the record, how likely the
model is to repeat a move hardly depends on how much it wanted the move
before. The valence-free record does nearly the same (slope
\PriorSlopeNeutCI), so the flattening comes with writing the move into the
transcript, not with the failure message. The abstract description, which
reports the same failure without the move string, keeps most of the
preference (\PriorSlopeAbstractCI).

Flattening has opposite consequences at the two ends of the prior
(Figure~\ref{fig:prior}, right).
\begin{itemize}[leftmargin=*,itemsep=2pt,topsep=2pt]
  \item Moves the model would not have played (prior below 1\%,
  \PriorBinLowN\ items) become more likely: $\Gain = \PriorBinLowGCI$ nats
  per token. Greedy decoding never produced them before the attempt
  (\PriorBinLowGreedyPre); after the verbatim record it produces them on
  \PriorBinLowGreedyPost\ of items, after the abstract description on
  \PriorBinLowGreedyAbstract.
  \item Moves the model was likely to play (prior above 30\%,
  \PriorBinHighN\ items) become less likely: $\Gain = \PriorBinHighGCI$.
  Greedy repetition falls from \PriorBinHighGreedyPre\ to
  \PriorBinHighGreedyPost\ after the verbatim record, but only to
  \PriorBinHighGreedyAbstract\ after the abstract description.
\end{itemize}

The fixed items of the original design lean towards the first kind. A
perturbed or random failing move is usually one the model would not propose:
\PriorSharedLowShare\ of the shared items have a prior below 1\% and
\PriorSharedHighShare\ one above 30\%, against \PriorModelLowShare\ and
\PriorModelHighShare\ of the model-proposed items. That is why the fixed items
invert. A failure in free
play is of the second kind by construction, since the agent wrote the move
because it was likely. On the model-proposed items, greedy decoding repeats
the failed move on \PriorModelGreedyPost\ of items after the verbatim record,
against \PriorModelGreedyPre\ before the attempt and
\PriorModelGreedyAbstract\ after the abstract description.

Two qualifications. First, $\Gain$ contains the prior, so a slope below one
is the same statement as $\Gain$ growing with the prior, and any change of
context unrelated to the move would weaken the dependence on the prior to
some degree. The informative comparison is therefore between records that
differ in one thing: writing the move flattens the preference, describing the
failure without it does not. Second, the flattening is format-dependent. In
the completion format of RQ2 the slope is larger (\PriorSlopeFailBaseCI\ for
the OLMo 2 base model), in RAGEN's format of RQ3 smaller
(\PriorSlopeFailRagenCI).

\section{RQ2: which training stage?}
\label{sec:rq2}

\begin{table}[t]
\centering
\scriptsize
\setlength{\tabcolsep}{2.5pt}
\begin{tabular}{l c c c c c c}
\toprule
& \multicolumn{3}{c}{$G$/tok, completion format} & \multicolumn{2}{c}{mean over notations} & chat \\
\cmidrule(lr){2-4}\cmidrule(lr){5-6}\cmidrule(lr){7-7}
Stage & short & verbose & index & $\Delta_{\mathrm{copy}}$/tok & $\Delta_{\mathrm{sem}}$/tok & $G$/tok (short) \\
\midrule
base & \textbf{\textminus 0.76\,\ci{\textminus 1.07}{\textminus 0.48}} & \textbf{\textminus 0.20\,\ci{\textminus 0.29}{\textminus 0.12}} & \textbf{\textminus 0.59\,\ci{\textminus 0.81}{\textminus 0.38}} & 0.49 & 0.03 & -- \\
SFT & \textbf{\textminus 0.56\,\ci{\textminus 0.90}{\textminus 0.22}} & \textbf{\textminus 0.16\,\ci{\textminus 0.24}{\textminus 0.08}} & \textbf{\textminus 0.54\,\ci{\textminus 0.86}{\textminus 0.24}} & 0.28 & 0.14 & \textminus 0.30\,\ci{\textminus 0.72}{0.10} \\
DPO & \textminus 0.16\,\ci{\textminus 0.63}{0.27} & \textminus 0.06\,\ci{\textminus 0.16}{0.03} & \textminus 0.13\,\ci{\textminus 0.53}{0.25} & \textminus 0.23 & 0.35 & 0.16\,\ci{\textminus 0.32}{0.62} \\
RLVR1 & \textminus 0.24\,\ci{\textminus 0.71}{0.20} & \textminus 0.06\,\ci{\textminus 0.15}{0.03} & \textminus 0.15\,\ci{\textminus 0.54}{0.23} & \textminus 0.20 & 0.35 & \textminus 0.06\,\ci{\textminus 0.49}{0.35} \\
Instruct & \textminus 0.26\,\ci{\textminus 0.73}{0.19} & \textminus 0.06\,\ci{\textminus 0.15}{0.03} & \textminus 0.17\,\ci{\textminus 0.58}{0.24} & \textminus 0.20 & 0.37 & \textminus 0.16\,\ci{\textminus 0.60}{0.27} \\
\bottomrule
\end{tabular}
\caption{RQ2: the OLMo 2 1B post-training ladder. Every checkpoint is probed on the same items in one completion format; the post-trained checkpoints also in their chat template. Pooled over the three games; shared items; intervals clustered on position.}
\label{tab:rq2}
\end{table}

\begin{figure}[t]
  \centering
  \includegraphics[width=\linewidth]{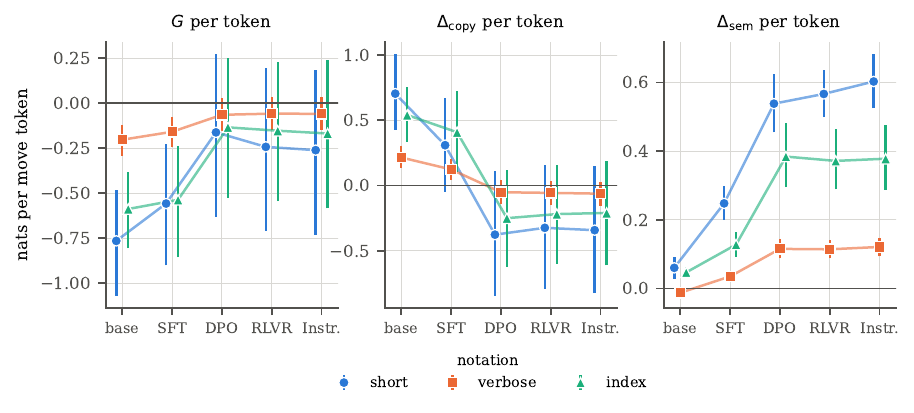}
  \caption{RQ2: the OLMo 2 1B post-training ladder in the completion format,
  by notation (shared items, nats per move token, 95\% intervals clustered on
  position).}
  \label{fig:rq2}
\end{figure}

The five OLMo 2 1B checkpoints differ only in post-training, and all are
probed on the same items in one completion format (Table~\ref{tab:rq2},
Figure~\ref{fig:rq2}). Each stage is compared with the one before it item by
item, so the contrasts below are paired over the same \RQtwoNItems\ (item,
notation) pairs from \RQtwoNPositions\ positions.

\paragraph{The base model already inverts.} Before any post-training, the
failure record raises the failed move's probability: $\Gain = \RQtwoGBaseCI$
nats per token, negative on \RQtwoFracNegBase\ of items. The surface form
carries it ($\copyterm = \RQtwoCopyBaseCI$); the failure message adds almost
nothing ($\semterm = \RQtwoSemBaseCI$). H4 holds: the inversion does not need
instruction tuning, which fits a copying mechanism learned in pretraining.

\paragraph{Preference tuning is where it shrinks.} Supervised fine-tuning
changes the gain little (paired change \RQtwoDGBaseSFTCI), although it already
lowers the copy term (\RQtwoDCopyBaseSFTCI). The DPO stage is the largest
step: the gain rises by \RQtwoDGSFTDPOCI\ nats per token, and at DPO it is
$\RQtwoGDPOCI$, an interval that includes zero. The copy term falls by
\RQtwoDCopySFTDPOCI\ and is no longer positive (\RQtwoCopyDPOCI): after DPO,
having written a move with a neutral acknowledgement no longer makes it more
likely. The semantic term moves the other way, by \RQtwoDSemSFTDPOCI, to
\RQtwoSemDPOCI. Relative to a neutral acknowledgement, being told that the move
failed now makes it \emph{more} likely, which keeps the gain from turning
clearly positive. RLVR (\RQtwoDGDPORLVRCI) and the final Instruct model
(\RQtwoDGRLVRInstructCI) change little after that. In their own chat template
the post-trained checkpoints show the same step from SFT to DPO
(\RQtwoChatDGSFTDPOCI).

\paragraph{Answer.} Within this ladder, the inversion comes from the base
model, and preference tuning, not supervised fine-tuning or RL with
verifiable rewards, is the stage that removes most of it. One ladder at one
scale cannot say whether this is a property of DPO or of this recipe's
preference data.

\section{Mechanism test: does the notation matter?}
\label{sec:mechanism}

\begin{table}[t]
\centering
\scriptsize
\setlength{\tabcolsep}{3pt}
\begin{tabular}{l c c c c }
\toprule
& RQ2 ladder & + prior & RQ1 models & + prior \\
\midrule
notation: verbose (vs.\ short) & {0.48}\,{\scriptsize(0.25)} & \textbf{0.53}\,{\scriptsize(0.15)} & {\textminus 0.16}\,{\scriptsize(0.78)} & \textbf{\textminus 2.12}\,{\scriptsize(0.34)} \\
notation: index (vs.\ short) & \textbf{\textminus 0.19}\,{\scriptsize(0.08)} & \textbf{\textminus 0.28}\,{\scriptsize(0.05)} & {\textminus 0.06}\,{\scriptsize(0.20)} & \textbf{\textminus 0.44}\,{\scriptsize(0.11)} \\
move length in tokens (z) & \textbf{\textminus 0.36}\,{\scriptsize(0.11)} & \textbf{\textminus 0.84}\,{\scriptsize(0.06)} & {\textminus 0.03}\,{\scriptsize(0.28)} & {\textminus 0.03}\,{\scriptsize(0.13)} \\
token overlap with a legal move & \textbf{0.86}\,{\scriptsize(0.17)} & \textbf{\textminus 0.55}\,{\scriptsize(0.11)} & \textbf{1.12}\,{\scriptsize(0.36)} & {0.05}\,{\scriptsize(0.15)} \\
prior log-prob.\ of the move, per token &  & \textbf{0.81}\,{\scriptsize(0.02)} &  & \textbf{0.92}\,{\scriptsize(0.01)} \\
losing (vs.\ illegal) &  & {\textminus 0.37}\,{\scriptsize(0.24)} &  & {\textminus 0.17}\,{\scriptsize(0.13)} \\
items & 1140 & 1140 & 1320 & 1320 \\
\bottomrule
\end{tabular}
\caption{Mechanism test: mixed-effects regression of $G$ per move token on notation, move length and token overlap, with random effects for position and model (shared items). Coefficients in nats per token with standard errors; bold: $p<0.05$. Positive coefficients mean less inversion. The ``+ prior'' columns add the failed move's log-probability before the attempt, which also enters $G$ itself (see text).}
\label{tab:mechanism}
\end{table}

If the inversion is carried by the move's surface form, the way a move is
written should change its size (H3: more inversion for longer moves and for
moves that share more tokens with the legal alternatives). Notation varies
within position in two places: the OLMo 2 ladder, where every item is probed
in all three notations, and two of the RQ1 models (SmolLM2-360M and
Qwen2.5-0.5B), which we also probed in the verbose and index notations.
Table~\ref{tab:mechanism} gives the mixed-effects regressions.

\paragraph{The ladder.} Longer moves invert more per token
(\MechRQtwoLength\ per standard deviation of length), in line with H3. The
index notation inverts slightly more than the short one (\MechRQtwoIndex), the
verbose one, if anything, less (\MechRQtwoVerbose). Token overlap with a legal move
goes against H3: moves that share more tokens with a legal alternative invert
\emph{less} (\MechRQtwoOverlap).

\paragraph{The chat models.} With notation varied on two models only,
neither notation contrast is distinguishable from zero (verbose
\MechRQoneVerbose, index \MechRQoneIndex) and neither is length
(\MechRQoneLength); overlap again goes against H3 (\MechRQoneOverlap).

\paragraph{Controlling for the prior.} Section~\ref{sec:prior} showed that the
failed move's prior largely sets the sign of $\Gain$. Notation, length and
overlap all change the prior, so we add it as a covariate (the ``+ prior''
columns). The prior then dominates (\MechRQtwoPriorPrior\ in the ladder,
\MechRQonePriorPrior\ in the chat models; it also enters $\Gain$ itself), and
several coefficients change sign: overlap now increases the inversion in the
ladder (\MechRQtwoPriorOverlap), as H3 predicted, and the verbose notation
inverts much more in the chat models (\MechRQonePriorVerbose).

\paragraph{Answer.} The notation changes the size of the inversion, and longer
moves invert more per token, but the direction of the other effects depends on
whether the prior is held fixed. Given equally plausible moves, overlap with a
legal move increases the inversion, as a copying account predicts; across
moves as they come, overlap goes with plausibility and the net effect is the
opposite. We read H3 as supported for length and only conditionally for
overlap.

\section{RQ3: the RAGEN format, before RL}
\label{sec:rq3}

\begin{table}[t]
\centering
\scriptsize
\setlength{\tabcolsep}{3pt}
\resizebox{\linewidth}{!}{%
\begin{tabular}{l l c c c c r}
\toprule
Model & Failure & $G$ & $\Delta_{\mathrm{copy}}$ & $\Delta_{\mathrm{sem}}$ & $G$, RAGEN's own feedback & positions \\
\midrule
Qwen2.5-0.5B & illegal (wall, box, unknown) & \textbf{\textminus 2.08\,\ci{\textminus 3.24}{\textminus 0.97}} & \textbf{2.05\,\ci{0.75}{3.36}} & 0.03\,\ci{\textminus 0.19}{0.28} & \textbf{\textminus 2.39\,\ci{\textminus 3.75}{\textminus 1.02}} & 45 \\
Qwen2.5-0.5B & deadlock push & 0.63\,\ci{\textminus 0.26}{1.47} & \textbf{\textminus 1.75\,\ci{\textminus 2.97}{\textminus 0.49}} & \textbf{1.12\,\ci{0.67}{1.56}} & -- & 15 \\
Qwen2.5-1.5B & illegal (wall, box, unknown) & \textbf{\textminus 3.82\,\ci{\textminus 5.54}{\textminus 2.22}} & \textbf{3.87\,\ci{2.29}{5.54}} & \textminus 0.05\,\ci{\textminus 0.13}{0.03} & \textbf{\textminus 3.89\,\ci{\textminus 5.52}{\textminus 2.33}} & 45 \\
Qwen2.5-1.5B & deadlock push & \textbf{0.50\,\ci{0.29}{0.75}} & \textminus 0.13\,\ci{\textminus 0.38}{0.14} & \textbf{\textminus 0.37\,\ci{\textminus 0.59}{\textminus 0.16}} & -- & 15 \\
\bottomrule
\end{tabular}}
\caption{RQ3: Sokoban in RAGEN's own prompt format (SimpleSokoban, no-think, one action per turn), RAGEN's two default models before any training. A RAGEN action is one token, so these are also per-token values. RAGEN's own feedback is its invalid-action warning, or nothing after a blocked move; it is defined for illegal moves only.}
\label{tab:rq3}
\end{table}

RQ3 asked whether RL on the game removes the inversion. We could measure only
its starting point: RAGEN's two default models, untrained, on Sokoban in
RAGEN's own prompt format, which is exactly what StarPO training would see
(Table~\ref{tab:rq3}).

\paragraph{Illegal actions invert strongly.} After an illegal action (a walk
into a wall, a push against a wall or box, or an unknown action), the record
makes that action more likely for both models: $\Gain = \RQthreeIllegalGQwenTwoSCI$
nats for Qwen2.5-0.5B and \RQthreeIllegalGQwenTwoLCI\ for Qwen2.5-1.5B
(\RQthreeIllegalNQwenTwoS\ positions each). An action is a single token here,
so these are also per-token values. The copy term carries it again
(\RQthreeIllegalCopyQwenTwoSCI\ and \RQthreeIllegalCopyQwenTwoLCI). RAGEN's own
feedback, its invalid-action warning or no message at all after a blocked
move, inverts at least as much as ours (\RQthreeNativeGQwenTwoSCI\ and
\RQthreeNativeGQwenTwoLCI). For Qwen2.5-0.5B the RAGEN format shows a clear
inversion where its chat template on Sokoban did not (\RQoneGQwenTwoSSokobanCI\
per token, on overlapping but not identical positions), and the flattening of
Section~\ref{sec:prior} is strongest in this format (slope
\PriorSlopeFailRagenCI).

\paragraph{Deadlock pushes do not.} When the push is legal but makes the room
unsolvable, the record lowers its probability: \RQthreeLosingGQwenTwoSCI\ for
Qwen2.5-0.5B, an interval that includes zero, and \RQthreeLosingGQwenTwoLCI\
for Qwen2.5-1.5B. These rest on only \RQthreeLosingNQwenTwoS\ positions.

\paragraph{Answer.} In the format that multi-turn RL on Sokoban trains in, the
untrained models start with a clear inversion on illegal actions, with
RAGEN's own feedback as with ours. Whether RL removes it, and whether that
transfers to games it was not trained on (H5), remains to be measured.

\section{RQ4: do the remedies transfer?}
\label{sec:rq4}

\begin{table}[t]
\centering
\scriptsize
\setlength{\tabcolsep}{3pt}
\begin{tabular}{l c c c c}
\toprule
Variant & Sokoban & Connect Four & Othello & pooled \\
\midrule
abstract description & \textbf{\textminus 1.18\,\ci{\textminus 1.65}{\textminus 0.74}} & \textminus 0.25\,\ci{\textminus 0.68}{0.17} & \textbf{\textminus 1.05\,\ci{\textminus 1.38}{\textminus 0.72}} & \textbf{\textminus 0.83\,\ci{\textminus 1.08}{\textminus 0.58}} \\
bare failure marker & \textbf{\textminus 0.95\,\ci{\textminus 1.45}{\textminus 0.48}} & \textminus 0.21\,\ci{\textminus 0.61}{0.18} & \textbf{\textminus 0.99\,\ci{\textminus 1.33}{\textminus 0.66}} & \textbf{\textminus 0.72\,\ci{\textminus 0.97}{\textminus 0.47}} \\
+ ``do not repeat'' & \textbf{0.11\,\ci{0.05}{0.17}} & 0.04\,\ci{\textminus 0.02}{0.10} & \textbf{\textminus 0.07\,\ci{\textminus 0.10}{\textminus 0.04}} & 0.03\,\ci{\textminus 0.00}{0.06} \\
message quotes the move & \textbf{0.49\,\ci{0.35}{0.63}} & \textbf{0.75\,\ci{0.67}{0.84}} & \textbf{1.09\,\ci{0.96}{1.22}} & \textbf{0.78\,\ci{0.70}{0.86}} \\
the move failed twice & \textbf{0.67\,\ci{0.53}{0.82}} & \textbf{1.71\,\ci{1.56}{1.85}} & \textbf{2.67\,\ci{2.48}{2.85}} & \textbf{1.68\,\ci{1.51}{1.86}} \\
a different move failed & \textbf{\textminus 1.10\,\ci{\textminus 1.55}{\textminus 0.68}} & \textbf{\textminus 0.45\,\ci{\textminus 0.88}{\textminus 0.02}} & \textbf{\textminus 0.77\,\ci{\textminus 1.11}{\textminus 0.45}} & \textbf{\textminus 0.77\,\ci{\textminus 1.02}{\textminus 0.54}} \\
\bottomrule
\end{tabular}
\caption{RQ4 in the probe: paired change in $\log P(a^{\times})$, nats, of each harness variant relative to the standard verbatim record, item by item (positive = repetition more likely). Pooled over the remedy-study models; intervals clustered on position.}
\label{tab:rq4}
\end{table}

\begin{figure}[t]
  \centering
  \includegraphics[width=\linewidth]{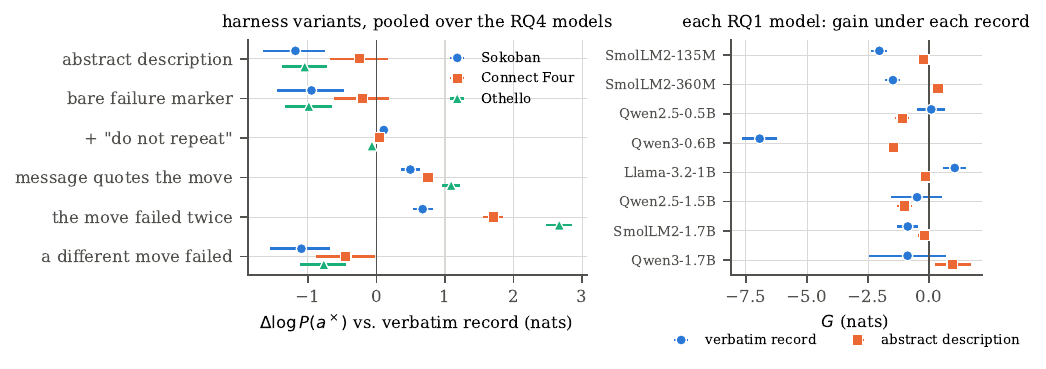}
  \caption{RQ4 in the probe. Left: change in $\log P(a^\times)$ of each
  harness variant against the verbatim record, item by item, pooled over the
  remedy-study models, by game. Right: for every RQ1 model, the gain under the
  verbatim record and under the abstract description (nats; shared items).
  95\% intervals clustered on position.}
  \label{fig:rq4}
\end{figure}

\paragraph{In the probe.} Each harness variant is compared with the verbatim
record item by item, as the change in the failed move's log-probability
(positive: repetition more likely), pooled over the \RQfourNumModels\ models of
the remedy study, \RQfourModelList\ (Table~\ref{tab:rq4},
Figure~\ref{fig:rq4}, left).
\begin{itemize}[leftmargin=*,itemsep=2pt,topsep=2pt]
  \item Replacing the move with the runtime's abstract description lowers the
  failed move's log-probability by \RQfourAbstractCI\ nats; a bare failure
  marker does about as well (\RQfourAbstractMinCI).
  \item Adding ``do not repeat'' to the system prompt changes nothing
  (\RQfourInstrCI), as in tool calling.
  \item Quoting the move in the failure message (\RQfourEchoCI) and recording
  it twice (\RQfourTwiceCI) make it more likely still, for every model in the
  remedy study.
  \item When a \emph{different} failing move is recorded, so that the scored
  move was never written, the scored move is less likely than when it was
  itself recorded (\RQfourOtherCI). What drives the repetition is the specific
  string in the transcript.
\end{itemize}

The abstract description is a core condition, so it is measured for all
\NumModelsRQone\ RQ1 models (Figure~\ref{fig:rq4}, right). Pooled, the gain is
\RQoneGTotalCI\ nats under the verbatim record and \RQoneGAbstractCI\ under the
abstract description, which removes \RQoneAbstractRemoved\ of the inversion,
close to the share the original reported for tool calling. It is not a fix
for every model: it lowers the failed move for \RQoneAbstractLowersN\ models
(\RQoneAbstractLowersList) and raises it for \RQoneAbstractRaisesList, models
whose copy term is negative (Table~\ref{tab:rq1}). Section~\ref{sec:prior} adds a second
caveat: for moves the model was likely to play, the abstract description stops
greedy repetition less well than the verbatim record does.

\paragraph{In free play.} \AgentModels\ played \AgentEpisodes\ greedy episodes
per game and harness against a random opponent (Figure~\ref{fig:agent};
Table~\ref{tab:agent} in the appendix). Both are weak players: \AgentIllegalOverall\ of all
moves are rejected and \AgentStepLimitShare\ of episodes end at the step
limit, so a win occurs in only \AgentCellsWithWins\ of the \AgentNumCells\
(model, game) combinations. Averaged over those combinations, the share of
failed moves that repeat an earlier failed move exactly is
\AgentVerbatimRepeat\ under the verbatim harness, \AgentInstrRepeat\ with the
instruction, \AgentAbstractRepeat\ with the abstract description and
\AgentDropRepeat\ with the clean retry, which, as in the original, is the worst
harness for repetition: greedy decoding in the restored context writes the same
move again. Here a failure is the model's own likely move, the case of
Section~\ref{sec:prior} in which the verbatim record lowers repetition rather
than raising it, which fits the verbatim harness repeating less than the clean
retry. Banning previously failed strings at the decoder removes exact repeats
(\AgentVerbatimBanRepeat\ and \AgentAbstractBanRepeat; at most
\AgentBanRepeatMax\ in any combination), but the share of illegal moves barely
moves (\AgentVerbatimIllegal\ under the verbatim harness, \AgentVerbatimBanIllegal\
with the ban): the models replace a repeated illegal move with a different
illegal one.

\begin{figure}[t]
  \centering
  \includegraphics[width=\linewidth]{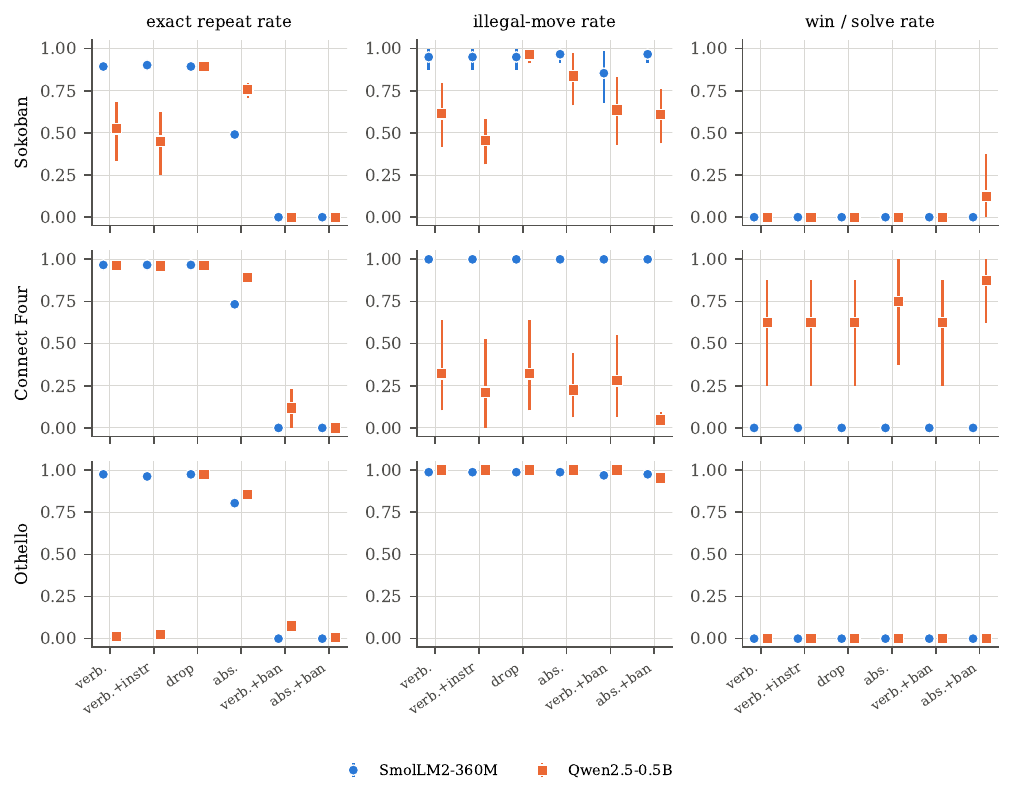}
  \caption{Free play under each harness (greedy decoding, random opponent).
  Points: mean over episodes; bars: 95\% bootstrap intervals over episodes.}
  \label{fig:agent}
\end{figure}

\paragraph{Answer.} The original's ranking of remedies transfers in part. The
instruction does nothing and the clean retry is the worst harness for
repetition, as before. The abstract description removes most of the inversion
on average, but not for every model and not for the moves a model is most
likely to repeat. Only the decoder ban reliably stops exact repetition, and at
these model sizes it does not turn repetition into legal play.

\section{Discussion}
\label{sec:discussion}

\paragraph{What the record does.} The simplest description of our results is
not ``the record makes the failed move more likely'' but ``the record makes
the failed move's probability stop depending on what the model thought of it
before''. Writing the move into the transcript pulls its probability towards
a level that is set by its being there. For a move the model would never have
chosen, that is an increase, and it is what the fixed probe items of the
original design measure. For the model's own likely mistake it is a decrease,
though an incomplete one. Both halves appear only when the move string is
written: the valence-free record flattens almost as much as the failure
record, and the abstract description, which carries the same information
without the string, does not. This is the copying account of the original
paper, seen from a different angle: what is copied is not simply ``more of the
failed move'' but the move's presence, which competes with the model's own
judgement of it.

\paragraph{What it means for the original's numbers.} The original probed tool
calls built by perturbing a correct call, the analogue of our shared items.
Our results suggest that such items mostly measure \emph{induced} repetition,
of actions the model would not otherwise have taken, and say less about the
repetition of a model's own likely errors. We did not test this in tool
calling. In games, the model's own likely errors are exactly what a free-play
failure is, and there the verbatim record lowers repetition relative to
deleting the attempt, but leaves much of it in place.

\paragraph{Where it comes from.} The OLMo 2 base model already inverts, and its
inversion is carried by the copy term, which points to copying mechanisms
learned in pretraining \citep{olsson2022induction} rather than to instruction
tuning. Preference tuning is the stage that removes most of it, and it does so
by removing the copy term's push (from \RQtwoCopyBaseCI\ at the base model to
\RQtwoCopyDPOCI\ after DPO), not by making the failure message more effective:
after DPO the message pushes slightly \emph{towards} the failed move relative
to a neutral acknowledgement. One recipe at one scale cannot tell
whether this is DPO itself or the content of its preference data, but it
suggests that the property under study is set early and shaped late.

\paragraph{For harness design.} Three recommendations survive from tool
calling: do not quote the failed move in the message (it makes repetition more
likely), do not rely on a ``do not repeat'' instruction, and do not retry from
the restored context under greedy decoding. One needs qualifying. The abstract
description protects against induced repetition, but for a move the model
strongly prefers it suppresses repetition less than the verbatim record does,
and for models that tend to avoid what is already in their context it can
backfire. In practice that argues for combining it with a constraint that does
not depend on the model's judgement, such as banning previously failed
strings at the decoder, which in our free play removed almost all exact
repetition. At these model sizes even that does not produce legal play.

\paragraph{For RL on games.} The default models of RAGEN start RL on Sokoban
with a clear inversion on illegal actions, under RAGEN's own feedback. An RL
run therefore begins in a regime where an illegal action, once recorded,
becomes more likely to be taken again. Whether training removes this, or
learns to work around it, is the measurement we could not make, and the one
we would make next.

\section{Limitations}
\label{sec:limitations}

\paragraph{Small models.} Every number comes from models between
\SmallestModelRQone\ and \LargestModelRQone\ parameters (and the 1B OLMo 2
ladder), run in float32 on one free-tier \GPUName. Float32 checkpoints larger
than about 3B do not fit in its memory, so the OLMo 3 7B ladder we had
planned as a scale check was not run, and we cannot say whether the
stage-wise picture of RQ2 holds at a larger scale.

\paragraph{No RL training.} Our proposal's RQ3 asked whether multi-turn RL on
Sokoban removes the inversion, with an SFT control on solver trajectories.
Both need GPU time well beyond a free notebook, so we report only the first
half: the inversion in RAGEN's prompt format before training. The training
configuration, the SFT-control data generator and the probes for trained
checkpoints are in the released code, untested on real runs.

\paragraph{Free play is near the floor.} The two models we could afford to
run in free play rarely play legal moves: \AgentIllegalOverall\ of their moves
are rejected, and \AgentStepLimitShare\ of episodes end at the step limit
rather than with a result. The harness comparison in free play therefore
mostly measures how illegal moves are distributed, not whether a harness wins
games. The teacher-forced probe, which does not depend on playing strength,
carries the main results.

\paragraph{The counterfactual success message is not fully coherent.} After
an illegal move the position does not change, so \mv{Move accepted.}\ followed
by the unchanged board is a slightly incoherent observation. The original
built success observations from the call's own arguments to avoid exactly
this; in a game, a coherent success would show a different position and change
the decision being scored. We therefore rely on the valence-free reference
(\mv{Move received.}) for the decomposition.

\paragraph{Losing is defined one move ahead.} A Connect Four move is
``losing'' only if the opponent can win immediately; deeper losses would need a
solver. This makes every label exact, at the cost of covering only the most
blatant losing moves. Sokoban deadlocks are exact at any depth, because the
rooms are small enough for exhaustive search. Losing moves are also the
smaller class: \RQoneLosingNPositions\ of the \RQoneNPositions\ positions in
RQ1.

\paragraph{Items are perturbations, not samples.} As in the original, the
primary items are built by operators so that every model sees byte-identical
items. Model-proposed items are the realistic case, but they differ across
models by construction and are reported separately.

\paragraph{Greedy decoding in free play.} Rollouts use greedy decoding, so
each (model, harness, game, episode) is one trajectory, and the clean-retry
harness is degenerate in the way the original described: an identical context
gives an identical move. With sampling it would not be.

\paragraph{Small samples.} The studies used the sample sizes planned for a
laptop CPU: 30 or 40 positions per game for each RQ1 model, 20 for each
stage of the OLMo 2 ladder, the remedy conditions on \RQfourNumModels\
models and \AgentEpisodes\ free-play episodes per cell
(Appendix~\ref{app:repro}). A second session on the full item pools was
prepared but not run. Intervals that include zero, such as those for losing
moves, the SFT stage and the ``do not repeat'' instruction, may therefore
reflect too few items rather than the absence of an effect.

\section{Conclusion}
\label{sec:conclusion}

Recording a failed move in an LLM player's transcript does not simply warn the
model off that move. It flattens the model's preference for it. Moves the
model would not have played become more likely, which reproduces the
feedback inversion of tool calling on fixed probe items, and moves it was
about to play become less likely, though greedy decoding still repeats a large
share of them. The effect is carried by the move string, is present before
any post-training, shrinks at the preference-tuning stage, and is clear in the
prompt format that RL on Sokoban starts from. Harnesses should keep the failed
string out of the message, should not rely on instructions or clean retries,
and should pair any rewording of the failure with a constraint at the decoder.

\section*{Data and code availability}

All code, the constructed probe items, the raw per-item log-probability
records, the free-play transcripts, the per-run environment metadata and the
research log are available at \RepoURL. Every number, table and figure in this
paper is regenerated from those records by one documented command; none is
transcribed by hand. The model checkpoints are public and pinned to the
revisions in Table~\ref{tab:models}.

\section*{Use of AI assistance}

The author used Anthropic's Claude, through the Claude Code command-line
interface, to write the experimental and analysis code and to draft and
revise the manuscript text. The author reviewed and edited the content and
takes full responsibility for it.

\bibliographystyle{plainnat}
\bibliography{references}

\clearpage
\appendix
\section{Reproduction}
\label{app:repro}

\paragraph{Code and environment.} Every number in this paper is produced by
code in the accompanying repository and written into the manuscript as a
macro; nothing is typed by hand, and a value that has not been measured
expands to a visible red ``TBD'' marker. The original paper's package is vendored
unchanged at commit \texttt{4af0401}, and a test checks every file against the
git blob hashes of that commit. Every reported run used one \GPUName\ (Google
Colab, free tier; CUDA \CudaVersion), Python \PythonVersion, PyTorch
\TorchVersion\ and Transformers \TransformersVersion, in float32 with TF32
\TFThirtyTwo. The reported results comprise \NumProbeRecords\ scored
(item, condition, candidate) records plus the free-play episodes.

\paragraph{Regression against the original.} The original paper's 100 ToolShed
items, run through this paper's probe runner on SmolLM2-135M on the original's
laptop (\CPUName, PyTorch \CPUTorchVersion, single-threaded), give
$\Gain = \RegressionG$, $\copyterm = \RegressionCopy$ and
$\semterm = \RegressionSem$ nats, the original's values, and reproduce every
stored log-probability to within \RegressionMaxDiff\ nats. A separate check
running the original code path on the same machine gives the same, which
separates environment drift from code differences. On the GPU the same items
give $\Gain = \RegressionGPUG$, with every log-probability within
\RegressionGPUMaxDiff\ nats of the CPU rows.

\paragraph{One GPU session.} Every reported study ran in one Colab session
with the sample sizes planned for a laptop CPU (configuration files
\texttt{cpu\_small.yaml} and \texttt{cpu\_large.yaml}; Section~\ref{sec:setup}
gives the positions per run), the remedy conditions on \RQfourNumModels\
models, and \AgentEpisodes\ free-play episodes per model, game and harness.
After that session we prepared a second one on the full item pools, with the
remedy conditions for every RQ1 model and 24 free-play episodes
(\texttt{gpu\_small.yaml}, \texttt{gpu\_large.yaml}). It was not run because
the free tier's GPU quota ran out, and no record from it enters this paper.
The research log records both decisions.

\paragraph{Keys and resumption.} Probe records are keyed by (model, revision,
format, game, notation, item, condition, candidate); rollouts by (model, game,
notation, format, harness, temperature, seed). Every run writes its git SHA, a
configuration hash, the library versions and the device.

\begin{table}[H]
\centering
\scriptsize
\begin{tabular}{l l l l}
\toprule
Model & Hugging Face id & Revision & Formats probed \\
\midrule
SmolLM2-135M & \texttt{HuggingFaceTB/SmolLM2-135M-Instruct} & \texttt{12fd25f773} & chat \\
SmolLM2-360M & \texttt{HuggingFaceTB/SmolLM2-360M-Instruct} & \texttt{a10cc1512e} & chat \\
SmolLM2-1.7B & \texttt{HuggingFaceTB/SmolLM2-1.7B-Instruct} & \texttt{31b70e2e86} & chat \\
Qwen2.5-0.5B & \texttt{Qwen/Qwen2.5-0.5B-Instruct} & \texttt{7ae557604a} & chat, ragen \\
Qwen2.5-1.5B & \texttt{Qwen/Qwen2.5-1.5B-Instruct} & \texttt{989aa7980e} & chat, ragen \\
Qwen3-0.6B & \texttt{Qwen/Qwen3-0.6B} & \texttt{c1899de289} & chat \\
Qwen3-1.7B & \texttt{Qwen/Qwen3-1.7B} & \texttt{70d244cc86} & chat \\
Llama-3.2-1B & \texttt{unsloth/Llama-3.2-1B-Instruct} & \texttt{5a8abab4a5} & chat \\
OLMo-2-1B base & \texttt{allenai/OLMo-2-0425-1B} & \texttt{a1847dff35} & completion \\
OLMo-2-1B SFT & \texttt{allenai/OLMo-2-0425-1B-SFT} & \texttt{0d85a3d037} & chat, completion \\
OLMo-2-1B DPO & \texttt{allenai/OLMo-2-0425-1B-DPO} & \texttt{c4b0485961} & chat, completion \\
OLMo-2-1B RLVR1 & \texttt{allenai/OLMo-2-0425-1B-RLVR1} & \texttt{12cb33498d} & chat, completion \\
OLMo-2-1B Instruct & \texttt{allenai/OLMo-2-0425-1B-Instruct} & \texttt{48d788eca8} & chat, completion \\
\bottomrule
\end{tabular}
\caption{Checkpoints and pinned revisions (first ten hex digits).}
\label{tab:models}
\end{table}

\section{Probe conditions in full}
\label{app:conditions}

\begin{table}[H]
\centering
\small
\begin{tabular}{l p{0.68\linewidth}}
\toprule
Condition & What the transcript records after the successful prefix \\
\midrule
\texttt{pre} & position s before the attempt (baseline; also the clean-retry context) \\
\texttt{fail} & attempt + failure message + s (standard harness, verbatim) \\
\texttt{succ} & attempt + counterfactual success message + s \\
\texttt{neut} & attempt + valence-free message + s \\
\texttt{abstract} & runtime-generated failure description + s, no verbatim move \\
\texttt{fail\_instr} & as fail, plus a do-not-repeat instruction in the system prompt \\
\texttt{fail\_echo} & as fail, but the message quotes the failed move \\
\texttt{fail\_k2} & the same move fails twice \\
\texttt{fail\_other} & a different failing move failed; the scored one was never written \\
\texttt{abstract\_min} & bare failure marker + s, no diagnosis and no verbatim move \\
\texttt{fail\_native} & RAGEN format only: RAGEN's own feedback (warning or nothing) + s \\
\bottomrule
\end{tabular}
\caption{Every probe condition. All conditions of an item share the system prompt, goal, successful prefix and position byte for byte; post-attempt conditions re-show the same position after their message.}
\label{tab:conditions}
\end{table}

\section{Messages}
\label{app:messages}

\begin{table}[H]
\centering
\scriptsize
\begin{tabular}{l l p{0.42\linewidth} p{0.25\linewidth}}
\toprule
Game & Failure type & Failure message (\texttt{fail}) & Gloss in the abstract description \\
\midrule
connect4 & \texttt{column\_full} & Illegal move: that column is already full. & the chosen column was already full \\
connect4 & \texttt{off\_board} & Illegal move: there is no such column. & the chosen location is not on the board \\
connect4 & \texttt{losing\_move} & That move lost: the opponent completes four in a row on the next turn. The position has been restored; choose again. & it let the opponent win on the next turn \\
othello & \texttt{occupied} & Illegal move: that square is already occupied. & the chosen square was already occupied \\
othello & \texttt{no\_flip} & Illegal move: a disc there would not flip any of the opponent's discs. & a disc there would not flip any discs \\
othello & \texttt{off\_board} & Illegal move: that square is not on the board. & the chosen location is not on the board \\
sokoban & \texttt{unknown\_action} & Unknown action: that move was not recognised. & the action was not recognised \\
sokoban & \texttt{blocked\_wall} & Blocked: a wall is in the way, so the player did not move. & a wall was in the way \\
sokoban & \texttt{blocked\_box} & Blocked: the box cannot be pushed that way, so nothing moved. & the box could not be pushed that way \\
sokoban & \texttt{deadlock\_push} & That push made the puzzle unsolvable: the box can no longer reach a target. The position has been restored; choose again. & the push made the puzzle unsolvable \\
\bottomrule
\end{tabular}
\caption{Observation messages (version 2). Valence-free: \texttt{Move received.}; counterfactual success: \texttt{Move accepted.}; abstract description: \texttt{[\{ordinal\} attempt failed: \{gloss\}; that move is not repeatable]}; bare marker: \texttt{[\{ordinal\} attempt failed]}; \texttt{fail\_echo} appends \texttt{ Your move was: \{move\}}; instruction: ``Important: some of your earlier moves failed. Do not repeat a move that has already failed; choose a different move.''.}
\label{tab:messages}
\end{table}

\section{Notations}
\label{app:notations}

\begin{table}[H]
\centering
\scriptsize
\begin{tabular}{l l >{\raggedright\arraybackslash}p{0.2\linewidth} >{\raggedright\arraybackslash}p{0.17\linewidth} >{\raggedright\arraybackslash}p{0.34\linewidth}}
\toprule
Game & Notation & Example & Invalid example & Rules text \\
\midrule
connect4 & short & \texttt{4} & \texttt{0} & Columns are numbered 1 to 7 from left to right. Reply with the column number only, for example: 4 \\
connect4 & verbose & \texttt{drop a disc into column 4} & \texttt{drop a disc into column 0} & Columns are numbered 1 to 7 from left to right. Reply with a phrase of the form: drop a disc into column 4 \\
connect4 & index & \texttt{3} & \texttt{-1} & Columns are indexed 0 to 6 from left to right. Reply with the column index only, for example: 3 \\
othello & short & \texttt{d3} & \texttt{a9} & Squares are named by column letter (a-h, left to right) and row number (1-8, top to bottom). Reply with the square only, for example: d3 \\
othello & verbose & \texttt{place a disc on column d, row 3} & \texttt{place a disc on column a, row 9} & Columns are lettered a-h from left to right and rows numbered 1-8 from top to bottom. Reply with a phrase of the form: place a disc on column d, row 3 \\
othello & index & \texttt{19} & \texttt{64} & Squares are indexed 00 to 63 row by row: index = 8 x (row - 1) + column position, with a=0 ... h=7. Reply with the two-digit index only, for example: 19 \\
sokoban & short & \texttt{U} & \texttt{N} & Moves are written U (up), D (down), L (left), R (right). Reply with one letter only, for example: U \\
sokoban & verbose & \texttt{move the player one square up} & \texttt{move the player one square north} & Reply with a phrase of the form: move the player one square up (or down, left, right) \\
sokoban & index & \texttt{0} & \texttt{4} & Moves are indexed 0 (up), 1 (down), 2 (left), 3 (right). Reply with the index only, for example: 0 \\
sokoban & ragen & \texttt{Up} & \texttt{North} & Up, Down, Left, Right \\
\bottomrule
\end{tabular}
\caption{Move notations. The RAGEN word form is used only in the RAGEN format.}
\label{tab:notations}
\end{table}

\section{Item statistics}
\label{app:items}

\begin{table}[H]
\centering
\small
\begin{tabular}{l l r}
\toprule
Game & Failure type & Decision points \\
\midrule
connect4 & \texttt{column\_full} & 22 \\
connect4 & \texttt{losing\_move} & 29 \\
connect4 & \texttt{off\_board} & 29 \\
othello & \texttt{no\_flip} & 27 \\
othello & \texttt{occupied} & 31 \\
othello & \texttt{off\_board} & 22 \\
sokoban & \texttt{blocked\_box} & 20 \\
sokoban & \texttt{blocked\_wall} & 24 \\
sokoban & \texttt{deadlock\_push} & 21 \\
sokoban & \texttt{unknown\_action} & 35 \\
\bottomrule
\end{tabular}
\caption{Item pools (seed 20261011; one decision point per position, each rendered in every notation). Sources: connect4: perturbation 60, perturbation+random 3, random 17; othello: perturbation 60, perturbation+random 1, random 19; sokoban: perturbation 80, perturbation+random 7, random 13.}
\label{tab:items}
\end{table}

\section{Free play in full}
\label{app:agent}

\begin{table}[t]
\centering
\scriptsize
\setlength{\tabcolsep}{3pt}
\begin{tabular}{l l l r r r r r}
\toprule
Model & Game & Harness & won/solved & exact repeat & illegal & wasted & loop \\
\midrule
qwen2.5-0.5b & Connect Four & verbatim & 62\% & 96\% & 32\% & 9.6 & 38\% \\
qwen2.5-0.5b & Connect Four & verbatim+instr & 62\% & 96\% & 21\% & 6.4 & 25\% \\
qwen2.5-0.5b & Connect Four & drop & 62\% & 96\% & 32\% & 9.6 & 38\% \\
qwen2.5-0.5b & Connect Four & abstract & 75\% & 89\% & 22\% & 5.9 & 25\% \\
qwen2.5-0.5b & Connect Four & verbatim+ban & 62\% & 12\% & 28\% & 8.4 & 25\% \\
qwen2.5-0.5b & Connect Four & abstract+ban & 88\% & 0\% & 5\% & 0.5 & 0\% \\
qwen2.5-0.5b & Othello & verbatim & 0\% & 1\% & 100\% & 40.0 & 50\% \\
qwen2.5-0.5b & Othello & verbatim+instr & 0\% & 2\% & 100\% & 40.0 & 100\% \\
qwen2.5-0.5b & Othello & drop & 0\% & 98\% & 100\% & 40.0 & 100\% \\
qwen2.5-0.5b & Othello & abstract & 0\% & 86\% & 100\% & 40.0 & 100\% \\
qwen2.5-0.5b & Othello & verbatim+ban & 0\% & 8\% & 100\% & 40.0 & 75\% \\
qwen2.5-0.5b & Othello & abstract+ban & 0\% & 1\% & 95\% & 38.1 & 12\% \\
qwen2.5-0.5b & Sokoban & verbatim & 0\% & 52\% & 61\% & 8.0 & 25\% \\
qwen2.5-0.5b & Sokoban & verbatim+instr & 0\% & 45\% & 45\% & 5.2 & 12\% \\
qwen2.5-0.5b & Sokoban & drop & 0\% & 90\% & 96\% & 9.6 & 0\% \\
qwen2.5-0.5b & Sokoban & abstract & 0\% & 76\% & 84\% & 8.5 & 0\% \\
qwen2.5-0.5b & Sokoban & verbatim+ban & 0\% & 0\% & 64\% & 9.2 & 0\% \\
qwen2.5-0.5b & Sokoban & abstract+ban & 12\% & 0\% & 61\% & 8.4 & 0\% \\
smollm2-360m & Connect Four & verbatim & 0\% & 97\% & 100\% & 30.0 & 100\% \\
smollm2-360m & Connect Four & verbatim+instr & 0\% & 97\% & 100\% & 30.0 & 100\% \\
smollm2-360m & Connect Four & drop & 0\% & 97\% & 100\% & 30.0 & 100\% \\
smollm2-360m & Connect Four & abstract & 0\% & 73\% & 100\% & 30.0 & 100\% \\
smollm2-360m & Connect Four & verbatim+ban & 0\% & 0\% & 100\% & 30.0 & 0\% \\
smollm2-360m & Connect Four & abstract+ban & 0\% & 0\% & 100\% & 30.0 & 0\% \\
smollm2-360m & Othello & verbatim & 0\% & 97\% & 99\% & 39.5 & 100\% \\
smollm2-360m & Othello & verbatim+instr & 0\% & 96\% & 99\% & 39.5 & 100\% \\
smollm2-360m & Othello & drop & 0\% & 97\% & 99\% & 39.5 & 100\% \\
smollm2-360m & Othello & abstract & 0\% & 80\% & 99\% & 39.5 & 100\% \\
smollm2-360m & Othello & verbatim+ban & 0\% & 0\% & 97\% & 38.8 & 0\% \\
smollm2-360m & Othello & abstract+ban & 0\% & 0\% & 98\% & 39.0 & 0\% \\
smollm2-360m & Sokoban & verbatim & 0\% & 89\% & 95\% & 9.5 & 0\% \\
smollm2-360m & Sokoban & verbatim+instr & 0\% & 90\% & 95\% & 10.8 & 25\% \\
smollm2-360m & Sokoban & drop & 0\% & 89\% & 95\% & 9.5 & 0\% \\
smollm2-360m & Sokoban & abstract & 0\% & 49\% & 97\% & 14.5 & 100\% \\
smollm2-360m & Sokoban & verbatim+ban & 0\% & 0\% & 85\% & 12.6 & 0\% \\
smollm2-360m & Sokoban & abstract+ban & 0\% & 0\% & 97\% & 14.5 & 0\% \\
\bottomrule
\end{tabular}
\caption{Free play under each harness (greedy decoding). Exact repeat: share of failed moves that repeat an earlier failed move byte for byte; illegal: share of all moves that the runtime rejected; wasted: rejected moves per episode; loop: episodes that hit the turn limit with a repeat.}
\label{tab:agent}
\end{table}

\end{document}